\documentclass[11pt]{article}

\usepackage[final]{acl}

\usepackage{times}
\usepackage{latexsym}
\usepackage[T1]{fontenc}
\usepackage[utf8]{inputenc}
\usepackage{microtype}
\usepackage{inconsolata}
\usepackage{graphicx}

\definecolor{darkblue}{HTML}{0E4D92}
\definecolor{oursrow}{HTML}{EEF3FA}

\usepackage{xurl}
\usepackage{booktabs}
\usepackage{float}
\usepackage{amsmath}
\usepackage{amsfonts}
\usepackage{array}
\usepackage{colortbl}
\usepackage{tabularx}
\usepackage{makecell}
\usepackage{tikz}
\usepackage{pgfplots}
\pgfplotsset{compat=1.18}
\usetikzlibrary{shapes.geometric, arrows.meta, positioning, calc, backgrounds, fit, shadows}
\usepackage[most]{tcolorbox}

\newcommand{\skillcode}[1]{\begingroup\urlstyle{tt}\nolinkurl{#1}\endgroup}

\newcommand{\skillpackage}{skill package}
\newcommand{\skillpackages}{skill packages}

\definecolor{appendixolive}{HTML}{8B8B5E}
\definecolor{appendixcream}{HTML}{FDFCF0}
\newcommand{\apdxhdr}[1]{\textcolor{white}{\textbf{#1}}}

\newcolumntype{A}[1]{>{\raggedright\arraybackslash}p{#1}}
\newcolumntype{Y}{>{\raggedright\arraybackslash}X}
\newcolumntype{Z}{>{\centering\arraybackslash}X}
\newcolumntype{C}[1]{>{\centering\arraybackslash}p{#1}}
\newcommand{\apdxstyle}{\setlength{\tabcolsep}{4pt}\renewcommand{\arraystretch}{1.18}}
\newcommand{\apdxband}{\rowcolor{appendixolive}[10pt][10pt]}

\newtcolorbox{ApdxTabFrame}[1][]{%
  enhanced,
  sharp corners,
  arc=0pt,
  boxrule=0.55pt,
  colframe=appendixolive,
  colback=appendixcream,
  left=6pt,
  right=6pt,
  top=4pt,
  bottom=4pt,
  width=\linewidth,
  #1,
}

\newtcblisting{GosPromptBox}[1]{%
  enhanced,
  sharp corners,
  arc=0pt,
  boxrule=0.55pt,
  colframe=appendixolive,
  colback=appendixcream,
  colbacktitle=appendixolive,
  coltitle=white,
  fonttitle=\bfseries\sffamily\footnotesize,
  titlerule=0mm,
  boxsep=3pt,
  left=8pt,
  right=8pt,
  top=2pt,
  bottom=6pt,
  title={#1},
  listing only,
  listing engine=listings,
  listing options={%
    basicstyle=\ttfamily\footnotesize,
    breaklines=true,
    columns=fullflexible,
    keepspaces=true,
    aboveskip=0pt,
    belowskip=0pt,
    xleftmargin=0pt,
  },
}

\newtcolorbox{ApdxCallout}[1][]{%
  enhanced,
  sharp corners,
  arc=0pt,
  boxrule=0.55pt,
  colframe=appendixolive,
  colback=appendixcream,
  fontupper=\small,
  left=8pt,
  right=8pt,
  top=6pt,
  bottom=6pt,
  #1,
}

\floatstyle{plain}
\newfloat{algorithm}{t}{loa}
\floatname{algorithm}{Algorithm}
\newenvironment{algobox}{%
  \begin{tcolorbox}[
    enhanced,
    sharp corners,
    arc=0pt,
    boxrule=0.55pt,
    colframe=appendixolive,
    colback=appendixcream,
    fontupper=\small,
    left=8pt,
    right=8pt,
    top=8pt,
    bottom=8pt,
  ]
}{%
  \end{tcolorbox}
}

\newcommand{\alginput}[1]{\textbf{Input:} #1\\}
\newcommand{\algoutput}[1]{\textbf{Output:} #1\\}
\newcommand{\algstep}[1]{#1\\}
\newcommand{\algindent}[1]{\hspace*{1.5em}#1\\}

\title{Graph-of-Skills: Dependency-Aware Structural Retrieval for Massive Agent Skills}

\author{Dawei Liu\textsuperscript{1}\thanks{Core Contribution.} \quad Zongxia Li\textsuperscript{2}\footnotemark[1] \quad Hongyang Du\textsuperscript{3} \quad Xiyang Wu\textsuperscript{2} \quad Shihang Gui\textsuperscript{3}\\ \textbf{Yongbei Kuang}\textsuperscript{4} \quad \textbf{Lichao Sun}\textsuperscript{5}\\[0.5ex] \textsuperscript{1}University of Pennsylvania \quad \textsuperscript{2}University of Maryland  \quad \textsuperscript{3}Brown University \quad \\\textsuperscript{4}Carnegie Mellon University  \quad \textsuperscript{5}Lehigh University \\ {\tt\small liudawei@seas.upenn.edu \quad zli12321@umd.edu \quad lis221@lehigh.edu}  }

\hypersetup{
  pdftitle={Graph-of-Skills: Dependency-Aware Structural Retrieval for Massive Agent Skills},
  pdfauthor={Dawei Liu, Zongxia Li, Hongyang Du, Xiyang Wu, Shihang Gui, Yongbei Kuang, Lichao Sun},
}

\begin{document}
\maketitle

\begin{abstract}
As LLM agents act across personal applications, web browsers, and other interfaces, their reusable skill libraries can scale to thousands of skills.
This scale introduces two challenges. First, loading the full library saturates the context window, driving up token costs, hallucination, and latency. Second, semantic retrieval surfaces topically relevant skills but can miss upstream and downstream prerequisite skills, creating a prerequisite gap that leaves the retrieved bundle insufficient for execution.
We present \textbf{Graph-of-Skills (GoS)}, an inference-time structural retrieval layer for large skill libraries.
GoS constructs an executable skill graph offline from skill packages, then retrieves a bounded, dependency-aware bundle through hybrid semantic--lexical seeding, reverse-aware Personalized PageRank, and context-budgeted hydration.
Across SkillsBench and ALFWorld, with three model families (Claude Sonnet 4.5, MiniMax M2.7, and GPT-5.2 Codex), GoS attains the highest average reward in all six model--benchmark blocks, at a fraction of the token cost of loading the full library.
On SkillsBench with GPT-5.2 Codex it raises average reward by $7.0$ absolute points over full skill loading, a $25.6\%$ relative gain, while cutting total tokens by $56.7\%$.
Ablations isolate the mechanism: replacing reverse traversal with forward propagation costs $9.1$ reward points, a larger loss than removing the graph altogether. The gain thus comes from traversing dependencies backwards, not from graph diffusion as such. A budget-matched retrieval study holding seeding, reranking, hydration, and context budget fixed reproduces the same ordering, with dependency-pair co-recovery falling from $0.654$ to $0.362$. Code is available at \url{https://github.com/davidliuk/graph-of-skills}.

\end{abstract}

\section{Introduction}
Large Language Model (LLM) agents solve complex technical tasks by invoking external tools, APIs, and reusable skills \citep{schick2023toolformer, mialon2023augmented}.
As tool and skill inventories grow from dozens to thousands or even tens of thousands of candidates \citep{patil2023gorilla, li2023apibank, xu2023toolbench, qin2024toolllm}, the core challenge shifts from deciding \emph{whether} to use a skill to retrieving the most relevant set of skills sufficient for a task.
\citet{shi2025toolret} show that retrieval is itself a major bottleneck in realistic tool ecosystems.

\begin{figure*}[!t]
    \centering
    \includegraphics[width=\textwidth]{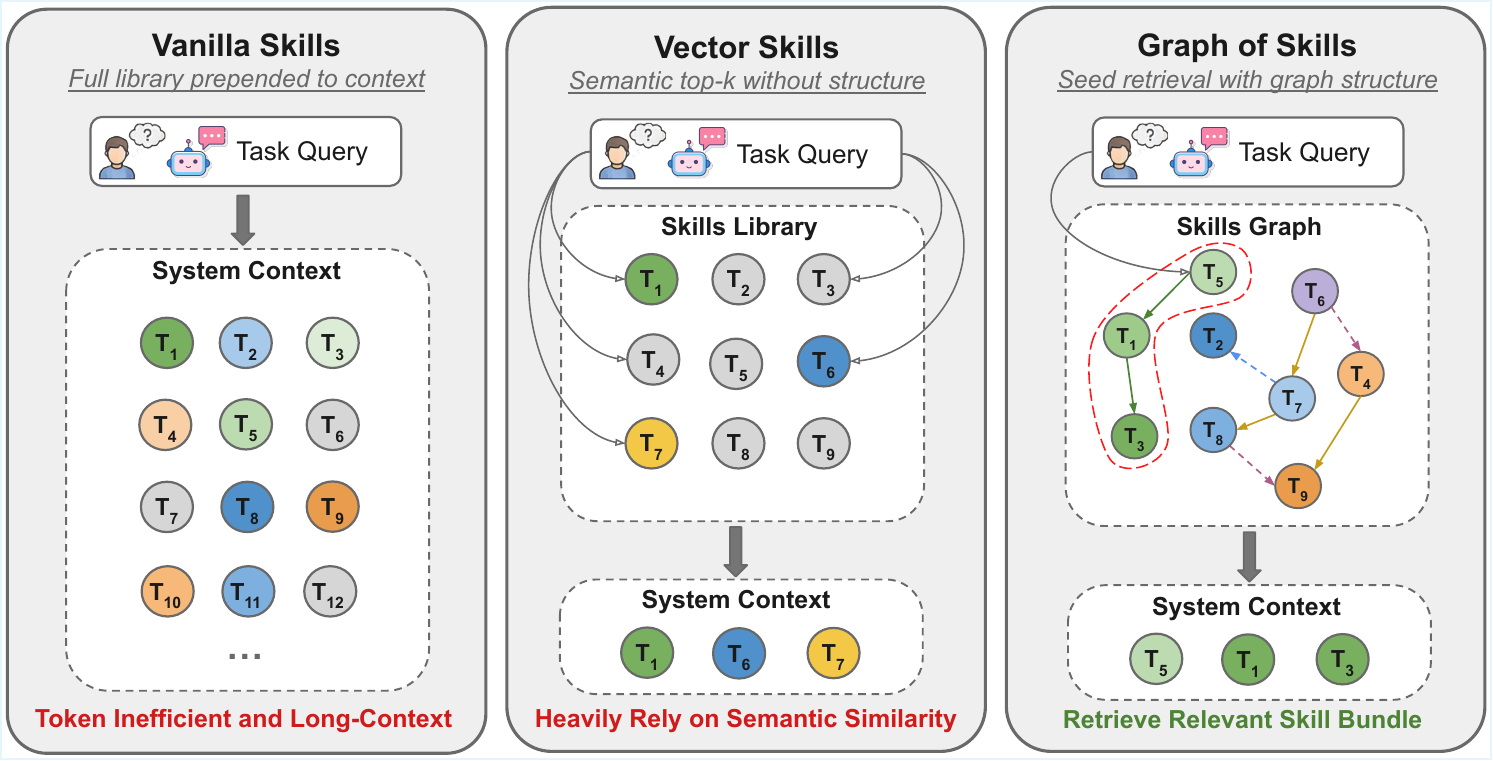}
    \caption{Conceptual comparison between flat skill loading, vector retrieval, and Graph-of-Skills (GoS). \textbf{Vanilla Skills} prepends the full skill library to the prompt, so relevant constraints and prerequisite skills become buried in an overloaded context. \textbf{Vector Skills} improves efficiency by returning semantically similar skills, but it can still miss a functionally required prerequisite outside the retrieved set, creating the prerequisite gap. \textbf{Graph-of-Skills} starts from hybrid semantic-lexical seeds and then performs structure-aware retrieval to recover prerequisite skills and assemble a compact execution bundle.}
    \label{fig:prerequisite_gap}
\end{figure*}

Two strategies dominate current practice.
\textbf{Vanilla Skills}~\citep{agentskills2026} prepends the entire skill set to the context window.
This can work for small toolsets, but it scales poorly: token cost grows linearly with library size, and critical domain constraints are easily overlooked in an overloaded context \citep{liu-etal-2024-lost}.
An alternative is \textbf{vector-based retrieval} \citep{lewis2020retrieval, wang2023voyager}, which improves efficiency by retrieving semantically similar skills.
However, semantic proximity does not imply executable sufficiency. In many engineering tasks, the top semantic match is a high-level solver, while the task also requires a lower-level parser, converter, setup utility, or preprocessor that is semantically weak but functionally necessary~\citep{qin2024toolllm, liu2025dynamic, patil2023gorilla} (see Figure~\ref{fig:prerequisite_gap}).

We present \textbf{Graph-of-Skills} (GoS), an inference-time structural retrieval layer for large local skill libraries that addresses both limitations. GoS constructs a directed multi-relational graph over local skill packages, where nodes are executable skills and edges encode prerequisite and workflow structure. At query time, GoS uses semantic and lexical signals only to identify a small seed set, then applies reverse-aware Personalized PageRank (PPR)~\citep{10.1145/511446.511513, 10471277} to recover additional skills that are structurally important for execution. The result is a bounded bundle that is relevant and more nearly dependency-complete than isolated top-$k$ retrieval.
Repository-scale skill infrastructures such as SkillNet and AgentSkillOS~\citep{skillnet2026, li2026agentskillos} address how skill ecosystems are created and organized. GoS targets the downstream question: what is the smallest subset of a local library sufficient to execute a task?

Our contributions are as follows:
\noindent\textbf{(1)} We introduce GoS, which combines offline typed-graph construction with inference-time structural retrieval to improve skill selection while cutting prompt cost.
 \textbf{(2)} We evaluate GoS across two benchmarks (SkillsBench, ALFWorld) and three model families. GoS attains the highest average reward in every model--benchmark block. Its largest margin comes on the 1{,}000-skill SkillsBench library under GPT-5.2 Codex, where it raises average reward by $7.0$ absolute points over full skill loading, a $25.6\%$ relative gain, while cutting total tokens by $56.7\%$. A library-size study from 200 to 2{,}000 skills shows the margin emerging once the library passes a few hundred skills. \textbf{(3)} We isolate the mechanism and find that traversal direction matters more than the graph itself: replacing reverse traversal with forward propagation costs $9.1$ reward points, a larger loss than removing the graph entirely. In a budget-matched retrieval study, forward propagation lowers dependency-pair co-recovery from $0.654$ to $0.362$.

\section{Related Work}

\paragraph{Tool Use, Discovery, and Retrieval for Agents.}
Early work on tool-augmented language models assumes small, fixed toolsets, where the challenge is deciding \emph{when} to invoke a tool and formatting the call~\citep{schick2023toolformer, mialon2023augmented}.
As tool sets grow from dozens to thousands~\citep{patil2023gorilla, li2023apibank, xu2023toolbench, qin2024toolllm}, agent tasks lengthen into multi-step terminal workflows~\citep{li2026long}, and context windows expand~\citep{singh2025openai, comanici2025gemini}, the problem shifts toward \emph{tool discovery} and \emph{tool retrieval}. Large tool universes therefore need scalable retrieval, and generic dense retrievers align poorly with real tool-use needs~\citep{shi2025toolret}.

\paragraph{Skill Acquisition and Self-Improvement.}
A parallel line asks how agents improve without curated supervision: by reusing verbalized experience~\citep{reflexion2023, zhao2024expel}, accumulating skills in embodied settings~\citep{wang2023voyager}, inducing programmatic skill sets~\citep{zheng2025skillweaver, wang2025asi, chen2026cuaskill}, or training on self-generated data and reward~\citep{li2026mm, li2025self}. ComfyClaw~\citep{li2026comfyclaw} introduces skill evolution to agentic visual workflow construction, distilling execution and verifier feedback into reusable Agent Skills. These methods grow capability; GoS assumes a fixed corpus and asks which subset to load.

\paragraph{Agent Skills Ecosystems.} Recent systems treat agent skills as reusable assets rather than ad hoc prompts~\citep{agentskills2026, skillnet2026, li2026agentskillos}.
SkillNet~\citep{skillnet2026} and AgentSkillOS~\citep{li2026agentskillos} advocate ecosystem-level categorization and chaining, while SkillsBench~\citep{li2026skillsbench} shows that curated skills help but that availability alone does not guarantee reliable use.
Registries such as SkillsMP, ClawHub, and LangSkills support packaging and search~\citep{skillsmp2026, clawhub2026, langskills2026}, but their interface remains search over individual skills or bundles.

\paragraph{Graph-Based Retrieval and Relational Memory.}
Graph-structured retrieval has improved knowledge access in document, memory, and tool-use settings, but its role differs across regimes. GraphRAG~\citep{edge2024graphrag} supports query-focused synthesis over documents, HippoRAG~\citep{gutierrez2024hipporag} models long-term memory as an associative graph, and ControlLLM~\citep{liu2023controlllm} and ToolNet~\citep{liu2024toolnet} place graph structure over tools, aiming at broad ecosystems rather than dependency-complete local bundles. These systems propagate along edges as stored; GoS weights reverse dependency transitions most heavily.
None of these studies examines retrieval over large local skill repositories: they target knowledge synthesis, relational QA, or graph-guided planning. Our setting instead needs an \emph{upstream} layer that, under a tight context budget, selects a small executable bundle \textit{before} generation, not a ranked list explored later.


\section{Methodology}
\label{sec:methodology}

\begin{figure*}[!t]
    \centering
    \includegraphics[width=1\linewidth]{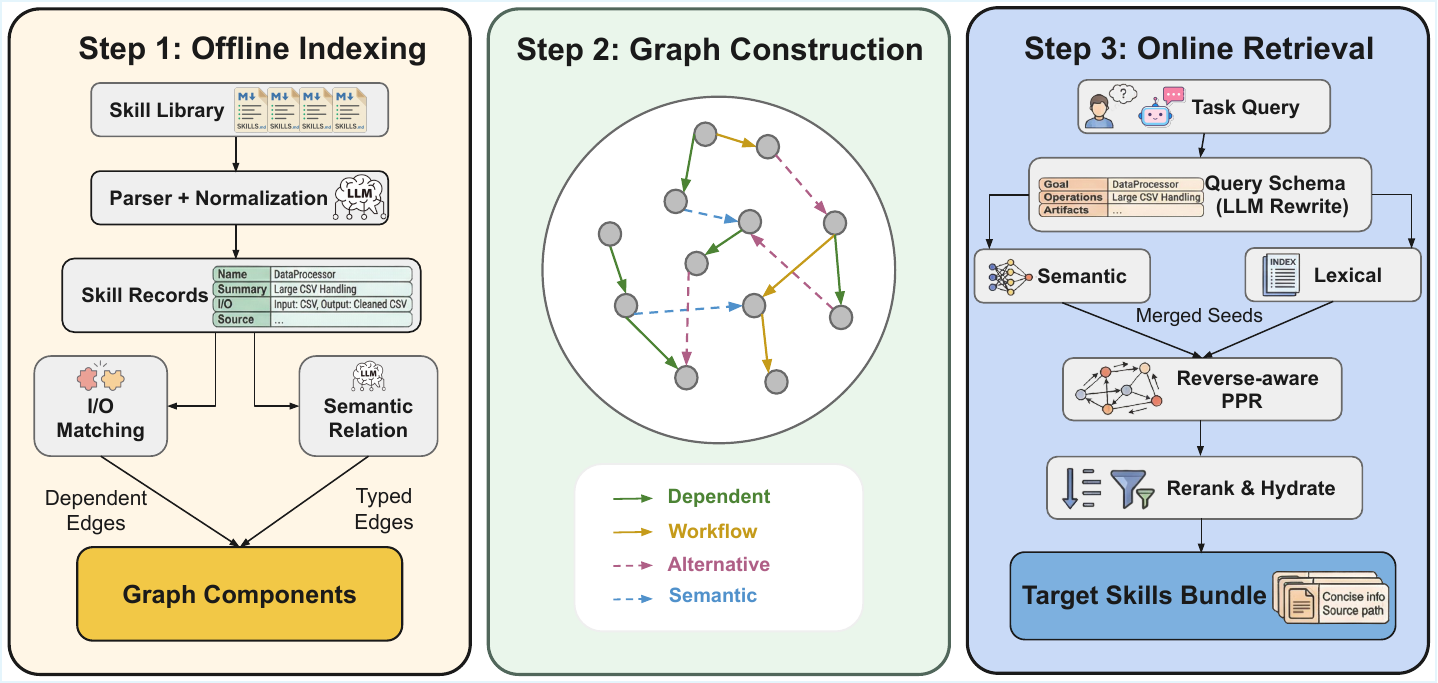}
    \caption{Overview of Graph-of-Skills (GoS). \textbf{Left:} offline indexing converts local \skillpackages{} into normalized skill records and typed edges. Dependency edges are induced from I/O compatibility, while workflow, semantic, and alternative relations are added through sparse validation. \textbf{Center:} the typed directed skill graph is the retrieval substrate; edge labels denote dependency, workflow, semantic, and alternative relations. \textbf{Right:} online retrieval maps a task query to a compact query schema, forms merged seeds from semantic and lexical retrieval, applies reverse-aware Personalized PageRank, and returns a budgeted execution bundle after reranking and hydration.}
    \label{fig:gos_arch_horizontal}
\end{figure*}

GoS is an inference-time retrieval layer for large local skill libraries (Figure~\ref{fig:gos_arch_horizontal}). It builds a typed graph offline from local \skillpackages{} and, at query time, returns a compact bundle that is relevant and more nearly dependency-complete than flat retrieval.

\subsection{Problem Setup}
Let $\mathcal{C} = \{d_1, \dots, d_m\}$ denote a local corpus of \skillpackages{}. Each package contains a primary specification document together with optional scripts, references, and auxiliary assets. GoS converts $\mathcal{C}$ into a typed directed graph
\begin{equation}
    G = (V, E, w, \phi),
\end{equation}

where each node $v \in V$ is a normalized executable skill record, each edge $e \in E$ connects two skills, $w(e) > 0$ is an edge weight, and $\phi(e) \in \mathcal{R}$ assigns an edge type from the relation set
\begin{equation}
\mathcal{R} = \{\mathrm{dep}, \mathrm{wf}, \mathrm{sem}, \mathrm{alt}\}.
\end{equation}

Given a task query $q$ and a context budget $\tau$, the retrieval problem is to return a bundle $B(q) \subseteq V$ that is simultaneously relevant, as execution-complete as possible, and compact. We formalize this as a budgeted selection problem,
\begin{equation}
\begin{aligned}
\max_{B \subseteq V} \; & \sum_{v \in B} \operatorname{rel}(v, q) + \beta \!\!\!\!\sum_{(u,v) \in E_{\mathrm{dep}}} \!\!\!\!\mathbb{I}[u \in B \wedge v \in B] \\
& \text{s.t.} \quad \operatorname{cost}(B) \le \tau
\end{aligned}
\label{eq:objective}
\end{equation}
where the first term favors query relevance, the second rewards dependency-complete bundles, and $\operatorname{cost}(B)$ is the prompt budget consumed by the hydrated bundle. This quadratic knapsack problem~\citep{pisinger2007quadratic} is NP-hard, so GoS approximates it in three stages: hybrid seed retrieval, reverse-aware diffusion, and budgeted reranking and hydration.

\subsection{Offline Graph Construction}
\paragraph{Skill Normalization.}
Each \skillpackage{} is parsed into a normalized skill record containing a canonical name, capability summary, I/O fields, domain tags, tooling, entrypoints, compatibility notes, and a stable local source path. Normalization is primarily deterministic: the system extracts executable fields wherever it can and calls a lightweight LLM pass only for retrieval-critical semantic fields that incomplete documentation leaves missing. Each skill thus becomes a retrieval unit an agent can consume directly at inference time.

\paragraph{Typed Relation Induction.}
GoS uses four edge types. \emph{Dependency} edges represent executable prerequisites and are the graph's primary structural relation. A dependency edge $v_i \rightarrow v_j$ is added deterministically, without an LLM call, whenever the I/O schemas of $v_i$ and $v_j$ overlap above a fixed threshold, so that $v_i$ can plausibly provide an artifact consumed by $v_j$. \emph{Workflow} edges capture multi-step pipelines, \emph{semantic} edges connect near-duplicate or topically adjacent skills, and \emph{alternative} edges link interchangeable strategies.

GoS builds the remaining relations, including further dependency edges, through sparse LLM validation. For each node it forms a small candidate pool from lexical similarity, semantic neighbors, and I/O-based expansion, and submits only the top-$k$ candidates to a constrained validator prompt (default $k{=}8$). Submitted candidate pairs therefore scale as $\mathcal{O}(N k)$ rather than $\mathcal{O}(N^2)$; because the candidates for one focus skill are batched into a single validator request, the number of calls is at most $N$. Incremental indexing further reduces this to $\mathcal{O}(|V_{\mathrm{new}}|\,N)$. This keeps construction tractable while anchoring the graph in executable structure rather than metadata proximity.

\subsection{Online Structural Retrieval}
\paragraph{Query Representation and Hybrid Seeding.}
Dense retrieval~\citep{karpukhin2020dense} is often effective at finding the visible top-level skill but weak at recovering semantically subtle prerequisites. Lexical retrieval~\citep{robertson2009probabilistic} is robust for concrete artifacts and filenames, but brittle under paraphrase. GoS combines both signals when seeding.

At retrieval time, the raw query is mapped to a lightweight retrieval schema containing the task goal, salient operations, referenced artifacts, and normalized keywords. An optional LLM rewrite produces this schema, following rewrite-then-retrieve pipelines~\citep{ma2023query}; without it, GoS falls back to deterministic lexical normalization. GoS then computes a semantic seed score $s_i^{\mathrm{sem}}(q)$ and a lexical seed score $s_i^{\mathrm{lex}}(q)$ for each candidate skill $v_i$, and merges them as
\begin{equation}
 z_i(q) = \eta \, s_i^{\mathrm{sem}}(q) + (1-\eta) \, s_i^{\mathrm{lex}}(q),
 \label{eq:seed}
\end{equation}
where $\eta \in [0,1]$ controls the semantic--lexical tradeoff. The initial seed distribution is obtained by normalizing the merged scores over the candidate pool,
\begin{equation}
\mathbf{p}_i = \frac{z_i(q)}{\sum_j z_j(q)}.
\end{equation}

\paragraph{Reverse-Aware Typed Diffusion.}
Let $A_r$ denote the weighted adjacency matrix for relation type $r \in \mathcal{R}$. To let retrieval move from a matched high-level skill toward likely prerequisites, GoS uses both forward and reverse transitions. For each relation type, we define a row-normalized forward operator $T_r^{\rightarrow}$ and a row-normalized reverse operator $T_r^{\leftarrow}$ obtained from $A_r$ and $A_r^{\top}$, respectively. GoS then forms the unified transition operator
\begin{equation}
T = \operatorname{RowNorm}\!\left( \sum_{r \in \mathcal{R}} \lambda_r \left(T_r^{\rightarrow} + \gamma_r T_r^{\leftarrow}\right) \right),
\label{eq:transition}
\end{equation}
where $\lambda_r \ge 0$ and $\sum_r \lambda_r = 1$ weight relation types, and $\gamma_r \ge 0$ controls how strongly reverse traversal is allowed for each type. In practice, reverse edges are inserted into the transition matrix using type-specific weights (see Table~\ref{tab:appendix_relation_weights}), and the matrix is row-normalized before diffusion.

The core retrieval step is a reverse-aware Personalized PageRank-style diffusion over this operator~\citep{page1999pagerank, jeh2003scaling, 10471277}:
\begin{equation}
\mathbf{s}^{(\ell+1)} = \alpha \, \mathbf{p} + (1-\alpha) T^{\top} \mathbf{s}^{(\ell)},
\label{eq:ppr}
\end{equation}
where $\alpha \in (0,1)$ is the restart parameter. Where flat top-$k$ retrieval scores each skill in isolation, diffusion spreads relevance across a local executable neighborhood. In particular, once a high-level solver is retrieved as a seed, upstream parser, setup, or preprocessing skills can still accumulate score through reverse dependency paths even when they are weak semantic matches to the query.

\paragraph{Budgeted Reranking and Hydration.}
The diffusion score alone is not enough: the final output must be compact and directly usable by an agent. GoS therefore reranks candidate skills by combining graph score with field-level query evidence:
\begin{equation}
\rho_i(q) = \mathbf{s}_i^{\star} + \mu \, m_i(q),
\label{eq:rerank}
\end{equation}
where $\mathbf{s}_i^{\star}$ is the converged diffusion score, $m_i(q)$ aggregates direct matches between the query and skill fields such as name, capability summary, artifacts, and entrypoints, and $\mu$ controls how much local grounding is preserved after graph expansion.

Candidates are hydrated in descending order of $\rho_i(q)$ under per-skill and global context budgets. Here, \emph{hydration} denotes materializing a selected skill into an agent-consumable payload: a stable source path, concise capability text, and the most relevant execution notes. The output is a bounded execution bundle that maximizes executable coverage within the prompt budget.


\section{Experiments}
We evaluate whether GoS improves agent performance and efficiency relative to flat full-library access and non-graph semantic retrieval.

\subsection{Experimental Setup}
We evaluate GoS on two benchmarks using the full released task sets.
\textbf{SkillsBench}~\citep{li2026skillsbench} contains a diverse set of real-world technical tasks across 11 domains, paired with curated Skills---structured packages of procedural knowledge (instructions, code templates, resources) that augment LLM agents at inference time.
The task domains span complex technical work such as macroeconomic detrending, power-grid feasibility analysis, 3D scan analysis, financial modeling, and seismic phase picking.
\textbf{ALFWorld}~\citep{shridhar2020alfworld} is an interactive simulator that aligns text descriptions and commands with a physically embodied robotic environment, built by combining TextWorld~\citep{cote2018textworld}, an engine for interactive text-based games, with the ALFRED dataset~\citep{shridhar2020alfred}.
Its tasks involve multi-step household activities such as navigating rooms, finding objects, and manipulating them.
ALFWorld is widely used in text-only mode as a benchmark for sequential decision making, where the agent reads room descriptions and issues commands to reach a goal~\citep{yao2023react, reflexion2023}; we evaluate all 140 episodes.

\paragraph{Baselines.}
We compare GoS against two baselines.
\textit{Vanilla Skills} exposes the entire skill library directly to the agent, maximizing recall but providing no retrieval-time compression. On ALFWorld, this follows the official Agent Skills format and reference repository~\citep{agentskills2026}.
\textit{Vector Skills} retrieves a bounded set by semantic similarity over the same embedding model GoS uses, \skillcode{openai/text-embedding-3-large}~\citep{openai2024textembedding3large} (3072 dimensions), isolating graph structure from the general benefit of retrieval-time compression.
\textit{GoS} uses the same embedding model as \textit{Vector Skills} but replaces flat nearest-neighbor retrieval with structure-aware retrieval over the skill graph. The critical comparison is therefore between flat semantic retrieval and dependency-aware structural retrieval under the same backbone and embedding setup, so embedding quality is not a confound.

\paragraph{Retrieval Budgets and Interface.}
Both retrieval methods return at most $N{=}8$ hydrated skills under a 2{,}400-character per-skill cap and a 12{,}000-character global cap, while \textit{Vanilla Skills} mounts the entire library. \textit{Vector Skills} and GoS share the same hydration format and stable \texttt{Source:} paths, and query rewriting is disabled throughout. System prompts differ only where \textit{Vanilla Skills} reaches the library directly rather than through a retrieval call. Table~\ref{results-table} is therefore an \emph{end-to-end system comparison}; Section~\ref{sec:matched-retrieval} isolates the graph operation itself under a fixed retrieval contract.

\paragraph{Models and Evaluation.}
Experiments are conducted with Claude Sonnet 4.5~\citep{anthropic2025claude45}, MiniMax M2.7~\citep{minimax2026m27}, and GPT-5.2 Codex~\citep{openai2025gpt52codex}.
Each model--method setting is run twice and we report the mean. Our primary metric is average reward; on ALFWorld rewards are binary, so it equals success rate. We also report average total tokens and agent-only runtime, measured from agent start to finish and excluding environment setup.

\paragraph{Evaluation Protocol.}
The same retry policy applies across the main and sensitivity experiments. If environment construction fails we rebuild and rerun up to twice more; tasks still failing are excluded as infrastructure failures rather than counted as model failures. A run that times out after substantive execution is kept and scored $0$; one that times out before starting is rerun.

\subsection{Main Results}

We present the main results in Table~\ref{results-table}.
Across all six model--benchmark blocks, GoS attains the highest average reward. It reduces average token usage relative to \textit{Vanilla Skills} in all six blocks and runtime in five, and beats \textit{Vector Skills} on reward everywhere while keeping prompt cost low.

\begin{table*}[t]
\caption{Main end-to-end results. \textbf{R}: average reward (\%), \textbf{T}: average total tokens, \textbf{S}: agent-only runtime (s); $\uparrow$/$\downarrow$ indicate whether larger or smaller is better. Results are means over two runs per setting. For ALFWorld, average reward equals success rate. Best results are in \textbf{bold}, second-best \underline{underlined}.}
\label{results-table}
\centering
\small
\setlength{\tabcolsep}{6pt}
\begin{tabular}{ll ccc c ccc}
\toprule
& & \multicolumn{3}{c}{SkillsBench} & & \multicolumn{3}{c}{ALFWorld} \\
\cmidrule(lr){3-5} \cmidrule(lr){7-9}
Model & Method & R $\uparrow$ & T $\downarrow$ & S $\downarrow$ & & R $\uparrow$ & T $\downarrow$ & S $\downarrow$ \\
\midrule
& Vanilla Skills & \underline{25.0} & 967,791 & 465.8 & & 89.3 & 1,524,401 & 53.2 \\
Claude Sonnet 4.5 & Vector Skills & 19.3 & \underline{894,640} & \textbf{357.3} & & \underline{93.6} & \underline{28,407} & \textbf{37.8} \\
\rowcolor{oursrow}\cellcolor{white} & GoS (ours) & \textbf{31.0} & \textbf{860,315} & \underline{364.9} & & \textbf{97.9} & \textbf{27,215} & \underline{49.2} \\
\midrule
& Vanilla Skills & \underline{17.2} & 942,113 & 580.7 & & 47.1 & 2,184,823 & 88.6 \\
MiniMax M2.7 & Vector Skills & 10.4 & \textbf{852,881} & \underline{552.9} & & \underline{50.7} & \underline{66,109} & \underline{73.4} \\
\rowcolor{oursrow}\cellcolor{white} & GoS (ours) & \textbf{18.7} & \underline{867,452} & \textbf{502.5} & & \textbf{54.3} & \textbf{65,227} & \textbf{68.8} \\
\midrule
& Vanilla Skills & \underline{27.4} & 3,187,749 & \textbf{686.8} & & 89.3 & 1,435,614 & 83.3 \\
GPT-5.2 Codex & Vector Skills & 21.5 & \textbf{1,243,648} & 773.0 & & \underline{92.9} & \textbf{34,436} & \textbf{57.0} \\
\rowcolor{oursrow}\cellcolor{white} & GoS (ours) & \textbf{34.4} & \underline{1,379,773} & \underline{715.6} & & \textbf{93.6} & \underline{46,462} & \underline{64.7} \\
\bottomrule
\end{tabular}
\end{table*}

\paragraph{Semantic retrieval struggles on long-horizon tasks.}
Many SkillsBench tasks are long-horizon and require combining relevant skills with prerequisites for environment setup, preprocessing, or output formatting---utilities that are rarely lexically salient in the task description. \textit{Vector Skills}, which ranks purely by embedding similarity to the query, often misses them, and on SkillsBench it falls \emph{below} flat exposure in all three blocks: from 25.0 to 19.3 under Claude Sonnet 4.5, 17.2 to 10.4 under MiniMax M2.7, and 27.4 to 21.5 under GPT-5.2 Codex. GoS improves on both baselines in all three blocks while substantially reducing token usage. What separates the two retrievers is not topical relevance but whether the returned bundle carries the prerequisites needed to finish the task.

\paragraph{The advantage transfers to embodied tasks.}
The same pattern largely holds on ALFWorld. Under Claude Sonnet 4.5, GoS reaches 97.9\% average success compared with 93.6\% for \textit{Vector Skills} and 89.3\% for \textit{Vanilla Skills}, cutting average total tokens from 1{,}524{,}401 to 27{,}215. Under MiniMax M2.7, GoS improves reward from 47.1\% and 50.7\% to 54.3\% with the lowest tokens and runtime in that block. Under GPT-5.2 Codex, the two retrievers differ by one episode (93.6\% vs. 92.9\%), and GoS uses more tokens and runtime than \textit{Vector Skills}; here the graph is a trade-off, not a win.

\paragraph{GoS offers the best efficiency--performance tradeoff.}
\textit{Vanilla Skills} keeps maximal recall but leaves the agent searching an unstructured set at growing cost, while \textit{Vector Skills} compresses that cost and returns sets that are often incomplete. Averaged over the three model families, GoS improves reward over \textit{Vector Skills} by 10.97 points on SkillsBench and 2.87 points on ALFWorld (Table~\ref{results-table}). With two runs per setting we do not test individual cells for significance; we rely on the pattern's consistency across six blocks and four library sizes rather than on any individual comparison.

\subsection{Qualitative Analysis}
Trajectories show where these aggregate differences come from. \skillcode{pedestrian-traffic-counting} needs a short but complete visual pipeline: extract frames, count pedestrians, emit the expected format. GoS surfaced exactly that pipeline---\skillcode{gemini-count-in-video}, \skillcode{video-frame-extraction}, \skillcode{openai-vision}---and scored $0.417$. \textit{Vanilla Skills} opened the same family of helpers, but only after searching the mounted library, and reached $0.267$; \textit{Vector Skills} retrieved topically related context that never became a workable plan ($0.041$). The converse case is instructive: on \skillcode{earthquake-phase-association}, \textit{Vanilla Skills} assembled a complete seismic stack and passed while GoS recovered only part of it and failed---structural retrieval helps only where the required neighborhood is present and connected. Across the ten cases in Appendix~\ref{sec:appendix_qualitative}, what distinguishes GoS is less the relevance of what it returns than how directly the bundle maps onto an executable decomposition of the task.

\section{Analysis}
\label{sec:ablation}
We now ask three questions: whether the advantage survives scale, which component produces it, and what the graph costs. The answers converge on traversal direction rather than on the presence of graph structure.

\subsection{Sensitivity to Skill Library Size}
We run an additional full-SkillsBench study with GPT-5.2 Codex while varying the library size from 200 to 500, 1{,}000, and 2{,}000 skills.
We report average reward, average total tokens, and agent-only runtime under the same retry and exclusion rules used in the main experiments in Figure~\ref{fig:scaleup-skillsbench}.

\begin{figure*}[t]
\centering
\begin{minipage}[c]{0.46\textwidth}
\centering
\footnotesize
\renewcommand{\arraystretch}{1.0} 
\setlength{\tabcolsep}{5pt}
\begin{tabular}{@{}llrrr@{}}
\toprule
N & Method & R $\uparrow$ & T (M) $\downarrow$ & S $\downarrow$ \\
\midrule
200  & Vanilla Skills    & \textbf{32.5} & 1.85 & \textbf{701.6} \\
     & Vector Skills & 21.2 & \textbf{1.06} & 833.8 \\
     \rowcolor{oursrow}\cellcolor{white} & GoS (ours)         & \underline{32.1} & \underline{1.36} & \underline{731.2} \\
\addlinespace[2pt]
500  & Vanilla Skills    & \underline{26.0} & 1.93 & \textbf{756.8} \\
     & Vector Skills & 20.7 & \textbf{1.10} & \underline{849.5} \\
     \rowcolor{oursrow}\cellcolor{white} & GoS (ours)         & \textbf{31.4} & \underline{1.16} & 890.3 \\
\addlinespace[2pt]
1000 & Vanilla Skills    & \underline{27.4} & 3.19 & \textbf{686.8} \\
     & Vector Skills & 21.5 & \textbf{1.24} & 773.0 \\
     \rowcolor{oursrow}\cellcolor{white} & GoS (ours)         & \textbf{34.4} & \underline{1.38} & \underline{715.6} \\
\addlinespace[2pt]
2000 & Vanilla Skills    & \underline{26.7} & 5.84 & \textbf{733.5} \\
     & Vector Skills & 23.8 & \textbf{1.11} & 799.8 \\
     \rowcolor{oursrow}\cellcolor{white} & GoS (ours)         & \textbf{31.3} & \underline{1.14} & \underline{788.0} \\
\bottomrule
\end{tabular}

\end{minipage}\hfill
\begin{minipage}[c]{0.52\textwidth}
\centering
\begin{tikzpicture}

\begin{axis}[
    name=plot1, 
    width=0.82\linewidth, 
    height=2.0cm,         
    scale only axis, 
    xmode=log,
    log basis x=10,
    xmin=180, xmax=2200, 
    xtick={200,500,1000,2000}, 
    xticklabels=\empty, 
    ymin=16, ymax=37, 
    ymajorgrids=true,
    grid style={draw=gray!15},
    axis line style={draw=black!50},
    tick style={draw=black!50},
    label style={font=\footnotesize},
    tick label style={font=\footnotesize},
    ylabel={Reward (\%)},
    legend columns=3, 
    legend style={
        draw=none, fill=none, font=\scriptsize, 
        at={(0.5,1.05)}, anchor=south, 
        /tikz/every even column/.append style={column sep=0.3cm}
    },
    legend cell align=left,
]
\addplot+[color=black!55, mark=square*, dashed, line width=0.95pt, mark size=2pt] coordinates {(200,32.5) (500,26.0) (1000,27.4) (2000,26.7)};
\addplot+[color=orange!85!black, mark=triangle*, dashdotted, line width=1.0pt, mark size=2.3pt] coordinates {(200,21.2) (500,20.7) (1000,21.5) (2000,23.8)};
\addplot+[color=darkblue, mark=*, line width=1.2pt, mark size=2.2pt] coordinates {(200,32.1) (500,31.4) (1000,34.4) (2000,31.3)};
\legend{Vanilla Skills, Vector Skills, GoS}
\end{axis}

\begin{axis}[
    name=plot2,
    at={(plot1.below south west)}, 
    anchor=north west,
    yshift=-0.15cm,       
    width=0.82\linewidth, 
    height=2.0cm,         
    scale only axis, 
    xmode=log,
    log basis x=10,
    xmin=180, xmax=2200, 
    xtick={200,500,1000,2000},
    xticklabels={200,500,1000,2000}, 
    ymin=0, ymax=6.5,
    ymajorgrids=true,
    grid style={draw=gray!15},
    axis line style={draw=black!50},
    tick style={draw=black!50},
    label style={font=\footnotesize},
    tick label style={font=\footnotesize},
    xlabel={Skill library size},
    ylabel={Tokens (M)}, 
]
\addplot+[color=black!55, mark=square*, dashed, line width=0.95pt, mark size=2pt] coordinates {(200,1.850) (500,1.925) (1000,3.188) (2000,5.836)};
\addplot+[color=orange!85!black, mark=triangle*, dashdotted, line width=1.0pt, mark size=2.3pt] coordinates {(200,1.063) (500,1.102) (1000,1.244) (2000,1.106)};
\addplot+[color=darkblue, mark=*, line width=1.2pt, mark size=2.2pt] coordinates {(200,1.340) (500,1.157) (1000,1.380) (2000,1.136)};
\end{axis}
\end{tikzpicture}
\end{minipage}

\caption{Sensitivity to library size on full SkillsBench under GPT-5.2 Codex. Left: compact summary table for Vanilla Skills, Vector Skills, and GoS, where \textbf{N} is library size and the remaining columns follow Table~\ref{results-table}, with tokens in millions. Right: reward and total-token trends as the skill repository grows from 200 to 2,000 skills. GoS maintains the highest reward once the library becomes moderately large, while both retrieval-based methods substantially slow the growth in prompt cost relative to flat exposure.}
\label{fig:scaleup-skillsbench}
\end{figure*}

\paragraph{Retrieval decouples prompt cost from library size.}
The strongest trend is in token cost. From 500 to 2{,}000 skills, \textit{Vanilla Skills} rises from 1.93M to 5.84M average total tokens, roughly $3\times$, while \textit{Vector Skills} stays near 1.10M--1.24M and GoS near 1.14M--1.38M. Retrieval breaks that coupling, and GoS does so without sacrificing reward.

\paragraph{The reward advantage appears once the library grows.}
At 200 skills \textit{Vanilla Skills} is still marginally ahead (32.5 vs. 32.1), but from 500 onward GoS leads both baselines by $5.4$, $7.0$, and $4.6$ absolute reward points. The margin peaks at 1{,}000 and remains substantial at 2{,}000, so a growing library does not weaken the benefit of dependency-aware retrieval.

\paragraph{Retrieval overhead is not the scaling bottleneck.}
Under GPT-5.2 Codex both retrieval methods are slower than \textit{Vanilla Skills} in agent-only runtime, reflecting the cost of searching before execution; the ordering reverses under Claude Sonnet 4.5 and MiniMax M2.7 (Table~\ref{results-table}). Either way, retrieval overhead is roughly constant in library size while the prompt cost of flat exposure grows with it, so the scaling burden lies in the prompt, not in retrieval, and only flat exposure incurs it.
\subsection{Component Ablation Study}
\begin{table}[t]
\caption{Component ablation on full SkillsBench with GPT-5.2 Codex and the 1{,}000-skill library. Columns as in Table~\ref{results-table}, with tokens in millions (uncached input + cached input + output).}
\label{tab:gos-ablation}
\centering
\footnotesize
\setlength{\tabcolsep}{4.5pt}
\begin{tabular}{lccc}
\toprule
Method & R $\uparrow$ & T (M) $\downarrow$ & S $\downarrow$ \\
\midrule
\rowcolor{oursrow} Full GoS (reverse-aware PPR) & \textbf{34.4} & 1.38 & \textbf{715.6} \\
Forward-only PPR ($\gamma_r{=}0$) & 25.3 & 1.20 & 742.7 \\
w/o graph propagation & 29.3 & \textbf{0.89} & 766.2 \\
w/o lexical + rerank & 26.7 & 1.01 & 747.7 \\
\bottomrule
\end{tabular}
\end{table}

We next ablate three components under the main SkillsBench configuration with GPT-5.2 Codex on the 1{,}000-skill library (Table~\ref{tab:gos-ablation}). The first sets $\gamma_r{=}0$ in Eq.~\eqref{eq:transition}, keeping the graph and the diffusion but removing every reverse transition; the second removes propagation, so the system cannot expand beyond its seeds; the third removes lexical retrieval and reranking, leaving semantic seeding alone.

\paragraph{Both graph propagation and lexical seeding contribute.}
Removing graph propagation cuts tokens from 1.38M to 0.89M but reward from 34.4 to 29.3 ($\downarrow5.1$); removing lexical retrieval and reranking cuts tokens to 1.01M and reward to 26.7 ($\downarrow7.7$). Between these two, seed quality matters more than expansion: weak initial skills give propagation less structure to exploit. Hybrid retrieval improves entry-point quality, and graph propagation converts those stronger seeds into a more execution-complete bundle.

\paragraph{Reverse traversal, not diffusion alone, drives the gain.}
Removing only the reverse transitions is the most damaging edit of the three: reward falls from 34.4 to 25.3 ($\downarrow9.1$), a larger drop than removing graph propagation altogether ($\downarrow5.1$), so forward-only propagation lands $4.0$ points \emph{below} the graph-free control. Because dependency edges are stored producer $\rightarrow$ consumer, forward diffusion spends mass on the downstream consumers of an already-matched skill rather than on the upstream prerequisites the bundle is missing, displacing useful seeds under a fixed budget. The effect is therefore specific to \emph{reverse-aware} propagation rather than to graph diffusion in general.

\subsection{Isolating the Graph Operation}
\label{sec:matched-retrieval}
\begin{table}[t]
\caption{Matched offline retrieval controls on the 200-skill SkillsBench library, averaged over three graph builds. All arms share the same 87 queries, five lexical seeds, reranker, top-8 selection, 2{,}400/12{,}000-character budgets, hydration, and source-path contract; only the graph operation changes. $R_{\mathrm{avail}}$/$R_{\mathrm{full}}$: available- and full-oracle recall; $C$: bundle completeness; $D$: dependency-pair co-recovery.}
\label{tab:matched-retrieval}
\centering
\footnotesize
\setlength{\tabcolsep}{4pt}
\begin{tabular}{lcccc}
\toprule
Retrieval operation & $R_{\mathrm{avail}}\uparrow$ & $R_{\mathrm{full}}\uparrow$ & $C\uparrow$ & $D\uparrow$ \\
\midrule
\rowcolor{oursrow} Reverse-aware PPR (GoS) & \textbf{0.575} & \textbf{0.502} & \textbf{0.421} & \textbf{0.654} \\
Forward-only PPR ($\gamma_r{=}0$) & 0.490 & 0.428 & 0.311 & 0.362 \\
No-Graph (matched flat) & \underline{0.542} & \underline{0.473} & \underline{0.368} & \underline{0.481} \\
One-Hop dependency & \underline{0.542} & \underline{0.473} & \underline{0.368} & \underline{0.481} \\
\bottomrule
\end{tabular}
\end{table}
The ablations above compare full agent runs, which entangle the graph operation with compact hydration, reranking, and the agent-facing interface. We therefore build a matched offline evaluator over the 200-skill library in which four retrieval operations---reverse-aware PPR, forward-only PPR, a graph-free flat retriever (\emph{No-Graph}), and deterministic incoming-dependency \emph{One-Hop} expansion---share the same 87 SkillsBench task queries, five lexical seeds, reranker, top-8 selection, character budgets, hydration, and source-path format. Realized context is matched as well, at 11{,}999.8/11{,}999.8/11{,}999.6/11{,}991.0 mean characters. Seeding is lexical throughout, removing embedding-service variation and leaving the graph operation as the only difference between arms.

\emph{Available-oracle recall} and \emph{exact completeness} score retrieval against the oracle skills present in the library, over the 76 of 87 queries that have at least one; \emph{full-oracle recall} uses all 87 and counts absent oracle skills as unretrieved; and \emph{dependency-pair co-recovery} is the fraction of induced dependency edges whose two available-oracle endpoints are retrieved together, a measure of structural rather than topical coherence. Appendix~\ref{sec:appendix_matched_retrieval} gives full definitions, per-build values, precision, and latency.

\paragraph{Direction matters more than the graph itself.}
Table~\ref{tab:matched-retrieval} places reverse-aware PPR ahead of every control on all four metrics. Averaging the three graph replicates within each task and bootstrapping over the 76 tasks, reverse minus forward available recall is $+0.085$ (95\% CI $[+0.044, +0.132]$), reverse minus No-Graph is $+0.033$ ($[+0.004, +0.066]$), and reverse minus One-Hop is $+0.033$ ($[+0.003, +0.065]$). Two conclusions follow. First, traversal direction is decisive: forward propagation is worse than not using the graph at all, reproducing at retrieval level the downstream ordering in Table~\ref{tab:gos-ablation}. Second, a budget-matched flat retriever is a strong baseline on recall, and the full graph's advantage over it on that metric is small. No-Graph and One-Hop coincide to three decimals: the two arms do return different bundles, differing in precision, size, and latency (Appendix~\ref{sec:appendix_matched_retrieval}), but a single undiffused hop over a graph of average degree $1.73$ with $59$ of $200$ nodes isolated reaches no oracle skill that flat retrieval missed. What helps is weighted traversal, not the presence of a dependency edge.

\paragraph{Recall counts skills; the prerequisite gap is about whether they compose.}
On the structural metric the arms separate far more sharply: dependency-pair co-recovery is 0.654 for reverse-aware PPR against 0.481 for flat and one-hop retrieval and 0.362 for forward propagation. This is what the prerequisite gap predicts, and it explains why a small recall margin here is consistent with the larger end-to-end margins in Table~\ref{results-table}. The two measure different things: this evaluator holds seeding fixed and isolates propagation alone, whereas Table~\ref{tab:gos-ablation} shows hybrid seeding carrying more of the downstream gain ($\downarrow7.7$ against $\downarrow5.1$). It is also offline and lexically seeded on the 200-skill library, so it bounds the mechanism rather than reproducing the deployed system. The oracle lists task-facing skills rather than a full prerequisite closure, so co-recovery measures structural coherence, not the recall of the induced edges. Bundle precision is essentially unchanged across the graph and flat arms ($0.280$ against $0.272$), and reverse propagation costs $10.89$\,ms at p50 against $9.32$\,ms, so the gains cost neither precision nor latency.

\paragraph{The LLM-validated relations carry the gain.}
The reverse-over-forward ordering holds on each of the three graphs separately---$0.612$ against $0.488$, $0.549$ against $0.475$, $0.563$ against $0.506$---so it is not an artifact of one favorable build. Restricting the graph to its deterministically induced I/O edges changes the picture: reverse traversal still improves completeness over forward propagation ($+0.053$, $[+0.013, +0.105]$) and raises co-recovery from $0.342$ to $0.500$, but its recall now matches rather than exceeds the flat retriever ($0.538$ against $0.542$). The LLM-validated relations therefore supply the coverage that turns the deterministic backbone into a retrieval gain. One caveat bounds all four arms equally: the 87 SkillsBench tasks name 195 distinct oracle skills, 34 of which are absent from every evaluated collection, so only 75 tasks have their full oracle present and absolute recall for every arm sits below its ceiling.

\subsection{Graph Quality, Cost, and Robustness}
\label{sec:graph-diagnostics}
\paragraph{The graph is sparse and not reproducible edge-for-edge.}
The 200-skill graph has 173 typed directed edges (73 deterministic I/O, 100 LLM-validated), average total degree 1.73, and 59 of 200 isolated nodes, so every margin we report is an average over a library where structural retrieval is inactive for a substantial minority of skills. A blinded 40-edge audit of the re-gated build finds every relation valid (40/40) but the best type in only 34/40: type boundaries, which set the diffusion weights, are the weak point of LLM-validated construction. Three relation rebuilds over a frozen node manifest give mean typed-directed edge Jaccard $0.607$, falling to $0.297$ when LLM node completion is repeated as well, so we version normalized manifests rather than regenerating them.

\paragraph{Cost is one-time and small; documentation quality is the main limiting factor.}
Construction costs \$0.29 at 200 skills and \$2.36 at 2{,}000, and online propagation adds about $1.5$\,ms at p50. Sweeping $\zeta$ from 0.4 to 0.8 varies the deterministic edge count from 172 to 46 while moving recall by less than $0.005$, so the threshold is not delicate. Stricter acceptance trades coverage for type precision rather than improving retrieval: replaying the same cached relation proposals through tighter gates removes 23 edges and lifts the isolate count from 59 to 73, while recall moves only from $0.575$ to $0.567$. Dropping half of all I/O fields costs $0.024$ recall, whereas dropping half of the descriptions costs $0.146$. ALFWorld packages carry descriptions but no formal I/O frontmatter, which is one reason its margins in Table~\ref{results-table} are narrower. Appendix~\ref{sec:appendix_graph_diagnostics} reports the full diagnostics.

\section{Conclusion}
Skill retrieval is a bottleneck for agents over massive libraries: vanilla loading is expensive and noisy, while vector retrieval misses prerequisite chains. GoS retrieves a small, jointly sufficient bundle instead---the target skill plus the parsers and preprocessors it needs.
It attains the highest average reward in all six model--benchmark blocks. Its strongest cell is the 1{,}000-skill SkillsBench under GPT-5.2 Codex, where it gains $7.0$ absolute points over full loading; on ALFWorld under the same model the margin narrows to an accuracy--efficiency trade-off. The advantage emerges once the library passes a few hundred skills and holds through 2{,}000. Ablations trace it to hybrid seeding and reverse traversal: forward-only propagation scores below a graph-free control.

\section*{Limitations}

\paragraph{Dependence on graph quality.} GoS is only as good as the graph it builds, and that graph is neither dense nor deterministic. On the 200-skill library 59 of 200 nodes are isolated, so for a substantial minority of skills structural retrieval reduces to flat retrieval, and thin or ambiguous documentation degrades edge quality further. Construction is also stochastic: rebuilds of the relation layer over a frozen node manifest agree only in part, and repeating LLM completion of the missing node fields lowers that agreement further. Reproducing a particular graph therefore means versioning normalized manifests rather than regenerating them, and libraries whose skills are sparsely described will benefit less than the ones we study.

\paragraph{Static and non-exhaustive relations.} Dependency, workflow, semantic, and alternative edges are a working taxonomy, not a complete ontology of skill relations. Conflict, deprecation, and dependencies conditioned on external tools or APIs are unrepresented, and negative or conditional relations would need propagation semantics beyond reverse-weighted diffusion. The graph is also built offline and never revised from execution traces or user feedback, so a wrong or missing relation persists, and diffusion can amplify an outdated skill by propagating relevance toward it. Provenance, confidence, conservative acceptance gates, versioned manifests, and incremental rebuilding limit that risk; learning edge corrections from failures and feedback remains future work.

\paragraph{Evaluation scope.} Our evidence covers two benchmarks, three model families, and libraries of 200 to 2{,}000 skills, none of which contains every oracle skill its tasks call for. The matched study in Section~\ref{sec:matched-retrieval} isolates the graph operation but stops at the prompt, while the end-to-end comparison in Table~\ref{results-table} measures agent reward without separating retrieval from the surrounding system; a prompt-identical downstream comparison would join the two. With no human-annotated prerequisite graph for SkillsBench, the recall of the induced edges is itself unmeasured. Larger libraries, an independently authored skill corpus, and multimodal or web-browsing agents built on vision-language models~\citep{li2025vlmsurvey}, whose notion of a skill differs from ours, all remain open.

\section*{Ethical Considerations}
This paper presents work whose goal is to advance the field of machine learning. We do not identify additional societal consequences that require separate discussion here.


\bibliography{custom}

\clearpage
\appendix

\setlength{\textfloatsep}{12pt plus 2pt minus 2pt}
\setlength{\dbltextfloatsep}{12pt plus 2pt minus 2pt}
\setlength{\floatsep}{10pt plus 2pt minus 2pt}
\setlength{\dblfloatsep}{10pt plus 2pt minus 2pt}
\setlength{\intextsep}{10pt plus 2pt minus 2pt}

\section{Appendix Overview}

To make the supplementary material easier to navigate, we briefly summarize the organization of the appendix before presenting the detailed evidence. The appendix is designed to complement the main paper along four axes: implementation fidelity, prompt/interface design, retrieval mechanics, and trajectory-grounded empirical analysis. Table~\ref{tab:appendix_roadmap} summarizes that organization.

\begin{table}[htbp]
\centering
\small
\caption{Appendix roadmap. The table summarizes the role of each supplementary section and how it complements the main paper.}
\label{tab:appendix_roadmap}
\begin{ApdxTabFrame}
\apdxstyle
\begin{tabularx}{\linewidth}{@{}A{0.28\linewidth}Y@{}}
\apdxband
\apdxhdr{Section} & \apdxhdr{Purpose} \\
\addlinespace[2.5pt]
Implementation Details & Documents how GoS is instantiated in code, including parsing, graph construction, hybrid seeding, diffusion, reranking, and hydration. \\
Prompt and Interface Examples & Shows representative internal prompts and agent-facing interface rules used during graph construction and online retrieval. \\
Core Retrieval Pseudocode & Gives compact pseudocode for the offline indexing pipeline and the online graph-based retrieval pipeline. \\
Matched Offline Retrieval Protocol & Specifies the budget-matched evaluator that isolates the graph operation, its metric definitions, per-build results, and the deterministic-core control. \\
Graph Quality, Cost, and Robustness & Reports graph statistics, the blinded edge-quality audit, construction stability and threshold sensitivity, offline cost, and metadata robustness. \\
Error Analysis & Separates retrieval misses, partial retrieval, execution drift, and infrastructure failures. \\
Qualitative Analysis & Provides trajectory-grounded case studies that compare the actual skill context exposed under different retrieval conditions. \\
\end{tabularx}
\end{ApdxTabFrame}
\end{table}

\section{Implementation Details}

This appendix summarizes the implementation decisions behind GoS and clarifies how the abstract pipeline in the main text is instantiated in code. The purpose is not to enumerate every engineering detail, but to expose the concrete design choices that determine graph quality, retrieval behavior, and the final agent-facing bundle.

\begin{table}[htbp]
\centering
\small
\caption{GoS implementation summary. The table highlights the main design choices that determine graph construction, retrieval, and agent-facing hydration.}
\label{tab:appendix_impl_summary}
\begin{ApdxTabFrame}
\apdxstyle
\begin{tabularx}{\linewidth}{@{}A{0.23\linewidth}Y@{}}
\apdxband
\apdxhdr{Component} & \apdxhdr{Implementation choice} \\
\addlinespace[2.5pt]
Node construction & Parser-first normalization from \texttt{SKILL.md} plus optional LLM completion of retrieval-critical semantic fields. \\
Dependency edges & Directed edges induced by bidirectional output--input compatibility checks. \\
Higher-order edges & Sparse LLM validation over a bounded candidate pool for workflow, semantic, and alternative relations. \\
Seed retrieval & Hybrid semantic and lexical seeding over normalized node fields. \\
Graph scoring & Reverse-aware typed Personalized PageRank with relation-specific reverse transitions. \\
Final output & Reranked, budgeted hydration into agent-usable skill payloads with stable \texttt{Source:} paths. \\
\end{tabularx}
\end{ApdxTabFrame}
\end{table}

\begin{table*}[!t]
\centering
\small
\caption{Normalized node fields used at retrieval time. The table lists the fields retained in each skill node and their retrieval role in GoS.}
\label{tab:appendix_node_fields}
\begin{ApdxTabFrame}
\apdxstyle
\begin{tabularx}{\linewidth}{@{}A{0.29\linewidth}A{0.19\linewidth}Y@{}}
\apdxband
\apdxhdr{Field} & \apdxhdr{Primary role} & \apdxhdr{Why it matters} \\
\addlinespace[2.5pt]
\skillcode{name}, \skillcode{description} & canonical identity and coarse semantic match & Provide the most stable high-level skill signature during lexical and semantic seeding. \\

\skillcode{one_line_capability} & concise capability abstraction & Helps retrieval align a task to what the skill actually does, rather than to document wording alone. \\

\skillcode{inputs}, \skillcode{outputs} & executable interface schema & Support deterministic dependency induction and help retrieval recover prerequisite producers and consumers. \\

\skillcode{domain_tags}, \skillcode{tooling} & technical context & Improve matching on domain-specific libraries, APIs, and workflows when task language is underspecified. \\

\skillcode{example_tasks} & usage priors & Improve recall for tasks described by objective or scenario rather than by direct tool names. \\

\skillcode{script_entrypoints} & implementation affordances & Help the agent discover reusable scripts instead of re-implementing logic from scratch. \\

\skillcode{compatibility}, \skillcode{allowed_tools} & execution constraints & Preserve operational restrictions that are important for verifier-aligned use. \\

\skillcode{source_path}, \skillcode{rendered_snippet} & hydration and agent consumption & Make retrieved skills directly inspectable inside the execution environment and keep the bundle compact. \\
\end{tabularx}
\end{ApdxTabFrame}
\end{table*}

\begin{table*}[!t]
\centering
\small
\caption{Relation types and edge weights used in reverse-aware graph diffusion. Forward weights govern transition mass along the stored edge direction; reverse weights govern backward propagation from a matched skill toward its likely prerequisites. Dependency edges, the only deterministically induced relation, receive the largest weight in both directions; the three LLM-validated relations receive smaller reverse weights to limit topical drift during diffusion.}
\label{tab:appendix_relation_weights}
\begin{ApdxTabFrame}
\apdxstyle
\begin{tabularx}{\linewidth}{@{}A{0.16\linewidth}A{0.10\linewidth}A{0.10\linewidth}A{0.22\linewidth}Y@{}}
\apdxband
\apdxhdr{Relation} & \apdxhdr{Forward} & \apdxhdr{Reverse} & \apdxhdr{Meaning} & \apdxhdr{Retrieval consequence} \\
\addlinespace[2.5pt]
Dependency & 1.0 & 1.0 & Skill $u$ produces an artifact consumed by skill $v$ & Strongest forward and backward propagation, since recovering prerequisites is the main purpose of GoS. \\

Workflow & 0.7 & 0.5 & Two skills are commonly chained in a concrete multi-step pipeline & Allows moderate backward expansion toward adjacent pipeline stages without dominating dependency evidence. \\

Semantic & 0.4 & 0.2 & Two skills belong to the same narrow capability cluster & Provides weak smoothing across near-neighbor skills while limiting topical drift. \\

Alternative & 0.3 & 0.1 & Two skills solve the same subproblem via different implementations & Provides minimal backward mass, mainly to keep interchangeable options reachable. \\
\end{tabularx}
\end{ApdxTabFrame}
\end{table*}

\begin{table*}[!t]
\centering
\small
\caption{GoS hyperparameters used in all reported experiments. Values are the defaults shipped in the released code and are held fixed across benchmarks, model families, and library sizes unless explicitly noted.}
\label{tab:appendix_hyperparams}
\begin{ApdxTabFrame}
\apdxstyle
\begin{tabularx}{\linewidth}{@{}A{0.30\linewidth}A{0.10\linewidth}Y@{}}
\apdxband
\apdxhdr{Hyperparameter} & \apdxhdr{Value} & \apdxhdr{Role} \\
\addlinespace[2.5pt]
PPR restart $\alpha$ & 0.2 & Teleport probability in Eq.~\eqref{eq:ppr}; controls how much mass stays near the seeds. \\
PPR max iterations & 50 & Power-iteration cap for the diffusion in Eq.~\eqref{eq:ppr}. \\
PPR tolerance & $10^{-6}$ & Convergence threshold on $\lVert \mathbf{s}^{(\ell+1)} - \mathbf{s}^{(\ell)} \rVert$. \\
Dependency I/O threshold $\zeta$ & 0.6 & Minimum schema-overlap score required to add a deterministic dependency edge. \\
Validation candidates per node $k$ & 8 & Top-$k$ candidates per node sent to the LLM relation validator. \\
Semantic seed pool & 20 & Dense-retrieval candidate count per query before merging. \\
Lexical seed pool & 20 & Lexical-retrieval candidate count per query before merging. \\
Seed set size & 5 & Number of merged seeds carried into graph diffusion. \\
Retrieval top-$N$ & 8 & Maximum number of hydrated skills returned to the agent. \\
Per-skill char budget & 2{,}400 & Truncation limit applied to each hydrated skill payload. \\
Global context char budget & 12{,}000 & Total character budget over the hydrated bundle. \\
\end{tabularx}
\end{ApdxTabFrame}
\end{table*}

\begin{ApdxCallout}
\paragraph{Pipeline Summary.}
Tables~\ref{tab:appendix_impl_summary}--\ref{tab:appendix_hyperparams} record the design choices, node fields, relation weights, and hyperparameters referred to below. GoS proceeds in two phases. Offline, it parses local skill packages into normalized nodes, adds dependency edges by I/O matching, and augments the graph with sparse workflow, semantic, and alternative relations. Online, it forms a hybrid semantic--lexical seed set, applies reverse-aware graph diffusion, and returns a reranked, budgeted bundle of execution-ready skills.
\end{ApdxCallout}

\paragraph{Implementation Substrate.}
GoS is implemented on top of a graph-backed retrieval substrate for workspace management, vector indexing, and graph storage, while replacing document-centric assumptions with skill-specific parsing, relation induction, and agent-oriented hydration. Concretely, the system maintains an HNSW vector index over skill representations together with a typed directed graph whose vertices are normalized skills and whose edges carry relation labels, directional semantics, and scalar weights. This yields a retrieval substrate in which semantic proximity and structural connectivity can be combined inside a single inference-time pipeline rather than treated as disjoint retrieval regimes.

\paragraph{Parser-First Skill Normalization.}
Each local \skillpackage{} is first parsed deterministically from its primary \texttt{SKILL.md} file and nearby package structure. The parser extracts the canonical name and description from YAML frontmatter, collects explicit input and output fields, recovers domain tags, tooling, example tasks, compatibility notes, and allowed tools from both frontmatter and markdown sections, and resolves script entrypoints by scanning the local \texttt{scripts/} directory when present. It also materializes a stable local source path and a rendered snippet used later for retrieval and hydration. This parser-first design keeps node construction anchored in package structure rather than relying on free-form semantic extraction.

\paragraph{LLM-Assisted Semantic Completion.}
When package documentation is incomplete, GoS optionally performs a lightweight LLM pass over the full markdown body to recover retrieval-critical semantic fields, including capability summaries, inputs, outputs, domain tags, tooling, and example tasks. Importantly, this stage is constrained to normalize a single skill node and is not used to emit graph relations directly. In other words, the LLM here serves as a high-precision semantic completion module for node attributes rather than as an unconstrained graph-construction oracle. The inferred fields are then merged back with the deterministic parse, favoring completed semantic fields only when they improve the retrieval representation.

\paragraph{Typed Node Representation.}
After normalization, each skill is serialized into a node record that stores both structured lists and compact textual views. Besides canonical descriptive fields, the node retains raw skill content, rendered snippets, script entrypoints, and a stable \texttt{Source:} path. This dual representation is operationally important: the graph and vector index operate over normalized fields, whereas the final agent-facing bundle requires concise but directly usable payloads that can be opened inside the execution environment without path reconstruction.

\paragraph{Directed Typed Relation Induction.}
The GoS graph is a typed directed graph rather than an undirected similarity graph. Dependency edges are induced deterministically by matching producer outputs against consumer inputs in both directions for each candidate pair. An edge $u \rightarrow v$ therefore has explicit executable semantics: $u$ can plausibly provide an artifact consumed by $v$. Since I/O compatibility is asymmetric, this dependency structure cannot be reduced to undirected similarity without losing prerequisite direction.

Non-dependency relations, namely \emph{workflow}, \emph{semantic}, and \emph{alternative}, are added through sparse LLM validation rather than dense all-pairs inference. For each node, GoS first forms a bounded candidate pool by combining lexical overlap, semantic neighbors from the vector index, and I/O-based candidate expansion. The LLM is then asked only to validate high-confidence relations inside this restricted pool. This two-stage design keeps graph construction tractable and biases the resulting graph toward precision rather than density.

\paragraph{Hybrid Seeding at Query Time.}
At retrieval time, GoS does not rely on vector search alone. The system first optionally rewrites the raw task request into a compact query schema containing the goal, operations, artifacts, constraints, and high-value keywords. Semantic seeding is then obtained from nearest-neighbor search in embedding space, while lexical seeding is computed from token overlap over normalized node fields such as name, capability, I/O descriptors, tooling, example tasks, entrypoints, and snippets. These candidate pools are merged and reranked before graph diffusion, so the graph is seeded by a hybrid entry set rather than by a single retriever. In practice, this detail matters because the quality of the initial seeds strongly influences whether later graph expansion can recover the correct prerequisite chain.

\paragraph{Reverse-Aware Structural Diffusion.}
Retrieval over the graph uses a Personalized PageRank-style diffusion operator constructed from the directed typed edges. The implementation first inserts forward transition mass along each stored edge, and then injects type-specific reverse transitions so that relevance can flow back from a matched high-level skill toward likely prerequisites. The reverse coefficients are largest for dependency edges and smaller for workflow, semantic, and alternative links, reflecting the fact that reverse traversal is most justified when recovering executable prerequisites. Operationally, this means the graph remains directed, but retrieval is explicitly reverse-aware. GoS therefore does not collapse the graph into an undirected graph; instead, it performs controlled backward propagation during scoring.

\paragraph{Reranking and Budgeted Hydration.}
The stationary graph score is not exposed to the agent directly. After diffusion, GoS reranks candidate skills by combining graph relevance with field-level query evidence, then hydrates only the top skills into an agent-facing bundle under both per-skill and global context budgets. Each hydrated payload includes a concise skill rendering, relevant execution notes, and the original local source path. The retrieval output therefore functions as a bounded execution context rather than a generic search-result list. This final budgeted hydration step is essential for preserving the efficiency advantage over flat all-skills loading while still presenting enough structure for downstream execution.

\paragraph{Section Summary.}
From an implementation perspective, GoS is best understood as a hybrid graph-construction and retrieval system: deterministic parsing and I/O matching provide a reliable executable backbone; optional LLM semantic completion improves node quality when documentation is incomplete; sparse LLM relation validation adds higher-order inter-skill structure; and reverse-aware graph diffusion converts a small hybrid seed set into a compact, more execution-complete bundle. These choices realize the central claim that retrieval must recover both the target skills and their prerequisite context.

\section{Prompt and Interface Examples}

\paragraph{Layered Prompt Design.}
GoS uses two prompt layers with deliberately separated responsibilities. The first layer operates \emph{inside} the indexing and retrieval stack, where LLMs are used only for constrained normalization, optional query rewriting, and sparse relation validation. The second layer operates at the \emph{agent interface}, where the environment prompt tells the downstream agent when to call retrieval, how to interpret the returned bundle, and how strongly to prefer reuse over open-ended exploration. This separation is methodologically important. Graph-side prompts determine what semantic structure enters the retrieval substrate, whereas agent-side prompts determine whether that retrieved structure is converted into a verifier-aligned execution plan.

\paragraph{Presentation Goal.}
We highlight only the prompt fragments and interface rules needed to understand the method, rather than enumerating full templates. Crucially, GoS does not rely on unconstrained prompt engineering. The internal prompts are used to normalize or validate bounded objects, and the external interface is used to constrain downstream behavior once retrieval has occurred. Together, these prompts form an interface contract between offline graph construction and online execution.

\begin{table*}[t]
\centering
\small
\caption{Representative prompt and interface components in GoS. The table highlights the small set of prompt contracts that shape graph construction and downstream agent behavior.}
\label{tab:appendix_prompt_examples}
\begin{ApdxTabFrame}[width=\textwidth]
\apdxstyle
\begin{tabularx}{\linewidth}{@{}A{0.19\textwidth}A{0.18\textwidth}A{0.24\textwidth}Y@{}}
\apdxband
\apdxhdr{Component} & \apdxhdr{Role} & \apdxhdr{Key constraint} & \apdxhdr{Intended effect} \\
\addlinespace[2.5pt]
\textbf{Internal Prompt A}\\Skill semantic completion & Normalize one skill document into retrieval-critical fields. & Preserve canonical \texttt{name}/\texttt{description}; fill only node-local semantic fields; return an empty \texttt{edges} list. & Improve node quality when \texttt{SKILL.md} is incomplete while preventing relation hallucination. \\

\textbf{Internal Prompt B}\\Relation validation & Verify whether a bounded candidate pair should receive a typed edge. & Restrict outputs to \{dependency, workflow, semantic, alternative\}; preserve exact source/target names; emit nothing when uncertain. & Keep the graph sparse and precise instead of generating dense all-pairs links. \\

\textbf{Internal Prompt C}\\Query rewrite & Rewrite a raw request into a compact retrieval schema. & Extract \texttt{goal}, \texttt{operations}, \texttt{artifacts}, \texttt{constraints}, and \texttt{keywords} without redefining the task. & Improve seed-stage lexical and semantic coverage while preserving task intent. \\

\textbf{Agent Interface}\\Retrieval usage contract & Tell the downstream agent when retrieval must be called and how the returned bundle should be used. & Run \texttt{graphskills-query} first; read \texttt{Retrieval Status}; use exact \texttt{Source:} paths; prefer adapting retrieved scripts; prioritize verifier-minimal behavior. & Make retrieval operational immediately, so the bundle narrows search instead of serving as optional background context. \\
\end{tabularx}
\end{ApdxTabFrame}
\end{table*}

\paragraph{Internal Prompt A: Skill Semantic Completion.}
Table~\ref{tab:appendix_prompt_examples} summarizes the four contracts. The semantic-completion prompt is intentionally narrow. It asks the model to normalize exactly one skill document and extract only retrieval-critical fields. In the implementation, the prompt explicitly preserves the canonical \texttt{name} and \texttt{description} when present, emphasizes high precision, and requires the returned \texttt{edges} list to be empty. This design reflects a conservative use of LLMs: the model is allowed to fill semantic gaps in node attributes, but not to invent graph structure. Operationally, this improves the quality of node representations used for indexing while avoiding a common failure mode in LLM-built graphs, namely relation over-generation. The prompt is therefore best understood as a constrained semantic completion module, not as a graph generator.

\paragraph{Internal Prompt B: Relation Validation.}
The relation-validation prompt is invoked only after GoS has formed a small candidate pool using lexical overlap, semantic neighbors, and I/O-based expansion. The prompt defines four edge types: \emph{dependency}, \emph{workflow}, \emph{semantic}, and \emph{alternative}. It also explicitly instructs the model to prefer sparse, high-precision edges, to emit nothing when uncertain, and to preserve exact skill names in the \texttt{source} and \texttt{target} fields. This makes the prompt function more like a relation verifier than a free-form graph generator. In practice, this design limits graph density and preserves the typed semantics used during reverse-aware diffusion.

\begin{figure}[t]
\begin{GosPromptBox}{Prompt excerpt: skill semantic completion}
1. Extract exactly one skill node from the document.
   ...
3. Infer only retrieval-critical fields: capability, inputs,
   outputs, domain_tags, tooling, example_tasks.
   ...
6. Use high precision. If uncertain, leave a field empty.
7. Do not invent relationships. Return an empty `edges` list.
\end{GosPromptBox}
\caption{Excerpt from Internal Prompt A. The model fills only node-local semantic fields and must return an empty \texttt{edges} list.}
\label{fig:prompt_semantic}
\end{figure}

Figure~\ref{fig:prompt_semantic} shows that GoS uses the LLM as a constrained semantic normalizer, not an unconstrained graph constructor. The output space is sharply restricted to node-local semantic completion.

\paragraph{Internal Prompt C: Query Rewrite.}
The optional query-rewrite prompt maps a raw task request to a compact retrieval schema with fields such as \texttt{goal}, \texttt{operations}, \texttt{artifacts}, \texttt{constraints}, and \texttt{keywords}. The prompt explicitly instructs the model not to invent benchmark-specific labels and to leave unclear fields empty. This is consistent with the retrieval objective in GoS: rewriting is used only to expose retrieval-critical technical terms such as file formats, APIs, protocols, and concrete operations. It is not intended to change the task itself. When rewriting is disabled or unavailable, the system falls back to deterministic lexical normalization, so query rewriting is a retrieval enhancement rather than a mandatory dependency. In other words, the prompt improves lexical and semantic coverage at the seed stage, but it is not allowed to redefine the problem.

\paragraph{Agent Interface Prompt.}
In the SkillsBench GoS environment, the agent is instructed to begin with a targeted retrieval query built from the task goal, artifact or format, operation or API, and verifier-critical constraints. The interface then forces the agent to read the retrieval status before continuing.

A \texttt{NO\_SKILL\_HIT} response means the agent must explicitly acknowledge that no relevant skill was found and proceed without claiming skill use. A \texttt{SKILL\_HIT} response means the returned bundle should be treated as a narrowing device: the agent is told to use the returned local source paths, inspect scripts before implementing from scratch, and prioritize the shortest path to verifier pass. This interface design matters because the main quality difference is often not whether some relevant skill exists somewhere in the library, but whether the agent receives a compact, execution-ready bundle early enough to affect the trajectory. In that sense, the interface prompt is part of the method rather than a presentation detail.

\begin{figure}[t]
\begin{GosPromptBox}{Prompt excerpt: agent retrieval interface}
Before writing any code, run:
graphskills-query "goal + artifact/format + operation/API +
verifier-critical constraint"

- If Retrieval Status: NO_SKILL_HIT, proceed without claiming skill use.
- If Retrieval Status: SKILL_HIT, use retrieved skills only as constraints.
- Use the exact Source path already returned.
- Prefer adapting retrieved scripts over broader re-implementation.
\end{GosPromptBox}
\caption{Excerpt from the agent interface prompt, which fixes when retrieval is called and how the returned bundle is used.}
\label{fig:prompt_interface}
\end{figure}

Figure~\ref{fig:prompt_interface} shows that the agent-facing interface is itself part of the method. It does not merely expose a search command. It constrains when retrieval is called, how the returned bundle is interpreted, and how aggressively the downstream agent is allowed to branch away from authoritative local implementations. Retrieval is therefore treated as an operational step: the bundle is meant to narrow the search space immediately, not merely provide background context.

\paragraph{Section Summary.}
Taken together, these examples show that prompt design in GoS is not generic scaffolding. The internal prompts constrain how semantic structure enters the graph; the external interface constrains how retrieved structure enters the agent's working context. The two layers therefore form an interface contract between offline graph construction and online agent execution. This contract is especially important in our setting because many failures are not pure retrieval misses, but retrieval-hit trajectories in which the agent still drifts unless the interface strongly biases it toward verifier-minimal reuse. This points to a broader methodological claim: in graph-augmented agent systems, retrieval quality depends not only on what is indexed, but also on how the retrieved structure is exposed and behaviorally enforced downstream.

\section{Core Retrieval Pseudocode}

\paragraph{Presentation Goal.}
For completeness, we provide pseudocode for the two main algorithmic stages of GoS: offline graph construction and online structural retrieval. The presentation is intentionally close to the implementation, but abstracted enough to foreground the method rather than the surrounding engineering details.

\begin{algorithm}[t]
\caption{Offline graph construction for GoS}
\label{alg:gos_offline}
\begin{algobox}
\alginput{Local skill documents $\mathcal{C}$, optional LLM services, linking budget $k$}
\algoutput{Typed directed graph $G=(V,E)$ and vector index over normalized skills}
\algstep{Initialize empty node set $V$ and edge set $E$.}
\algstep{For each skill document $d \in \mathcal{C}$:}
\algindent{Parse frontmatter and markdown from \texttt{SKILL.md}.}
\algindent{Extract deterministic fields (e.g., name, description, I/O, tags, tooling, and entrypoints).}
\algindent{If retrieval-critical semantic fields are incomplete, run constrained semantic completion to fill capability, inputs, outputs, and example tasks.}
\algindent{Serialize as skill node $v$ and add it to $V$.}
\algstep{For each ordered pair of nodes $(u,v)$ in a bounded candidate pool:}
\algindent{Compute producer--consumer overlap between outputs of $u$ and inputs of $v$.}
\algindent{If overlap exceeds threshold, add edge $u \rightarrow v$.}
\algstep{For each node $u$:}
\algindent{Form a sparse candidate set using lexical similarity, semantic neighbors, and I/O-based expansion.}
\algindent{Validate relations on this candidate set.}
\algindent{Add these validated edges to $E$.}
\algstep{Build a vector index over normalized nodes.}
\algstep{Persist the typed graph, vector index, and retrieval metadata to the GoS workspace.}
\end{algobox}
\end{algorithm}

\begin{algorithm}[t]
\caption{Online graph-based skill retrieval}
\label{alg:gos_online}
\begin{algobox}
\alginput{Query $q$, prebuilt GoS workspace, retrieval budget $\tau$}
\algoutput{Bounded execution bundle $B(q)$}
\algstep{Optionally rewrite $q$ into a compact schema containing goal, operations, artifacts, constraints, and keywords.}
\algstep{Retrieve semantic seeds from the vector index.}
\algstep{Retrieve lexical seeds from normalized fields.}
\algstep{Merge candidate pools to form seed distribution $\mathbf{p}$.}
\algstep{Build a typed transition matrix over the graph.}
\algindent{Add forward transition mass for each stored edge.}
\algindent{Add relation-specific reverse mass, with the largest reverse coefficient on dependency edges.}
\algstep{Run reverse-aware Personalized PageRank until convergence to obtain graph scores $\mathbf{s}^{\star}$.}
\algstep{Rerank candidate skills using graph score plus direct field-level evidence from the query.}
\algstep{Hydrate the ranked skills into agent-facing payloads with exact \texttt{Source:} paths and concise execution notes.}
\algstep{Truncate the bundle under per-skill and global budgets.}
\algstep{Return retrieval status, ranked bundle summary, bounded agent-facing context, and graph evidence among selected skills when it fits the budget.}
\end{algobox}
\end{algorithm}

\paragraph{Implementation Correspondence.}
Algorithm~\ref{alg:gos_offline} corresponds to the parser-first normalization and relation-induction logic in the GoS implementation. Algorithm~\ref{alg:gos_online} corresponds to the retrieval path and the reverse-aware Personalized PageRank utilities. In practice, these stages additionally include engineering details such as workspace bootstrapping, embedding-dimension checks, source-path rewriting for containerized environments, and context clipping under hard character budgets. We omit those details from the pseudocode because they are not conceptually central, but they are nevertheless important for stable end-to-end deployment.

\section{Matched Offline Retrieval Protocol}
\label{sec:appendix_matched_retrieval}

This appendix documents the matched retrieval evaluator summarized in Section~\ref{sec:matched-retrieval}. It isolates a question the end-to-end ablations leave open: whether the reward gains of GoS come from the graph operation itself or from the surrounding retrieval pipeline---hybrid seeding, reranking, budgeted hydration, and the agent-facing source-path contract.

\paragraph{Design.}
Four retrieval operations are evaluated on the 200-skill SkillsBench library: reverse-aware PPR as used by GoS, forward-only PPR obtained by setting $\gamma_r{=}0$ in Eq.~\eqref{eq:transition}, a graph-free flat retriever (\emph{No-Graph}), and deterministic incoming-dependency \emph{One-Hop} expansion that adds the direct dependency predecessors of each seed without any diffusion.

Every arm receives the same 87 SkillsBench task instructions as queries, the same five lexical seeds, the same reranker, the same top-$N{=}8$ selection, the same 2{,}400-character per-skill and 12{,}000-character global budgets, the same hydration routine and execution notes, and the same exact source-path format. Only the graph operation changes. Seeding is lexical throughout, which removes the embedding service from the comparison so that no arm benefits from embedding drift or API variation. The evaluator therefore measures the graph operation under a fixed lexical seed, not under the deployed hybrid-seed configuration.

Each arm is run against three independently constructed graphs (the fixed-node builds A, B, and C of Appendix~\ref{sec:appendix_graph_diagnostics}), so that graph-construction variance is carried through the comparison rather than absorbed into a single build.

\paragraph{Metric definitions.}
Let $O_t$ be the SkillsBench oracle skill set for task $t$, $A_t = O_t \cap V$ the subset present in the evaluated library, and $B_t$ the returned bundle.
\emph{Available-oracle recall} is $\frac{1}{|T_{+}|}\sum_{t \in T_{+}} |B_t \cap A_t| / |A_t|$, where $T_{+}$ is the set of tasks with $|A_t| > 0$.
\emph{Exact completeness} is the fraction of $t \in T_{+}$ with $A_t \subseteq B_t$.
\emph{Full-oracle recall} is computed over all 87 tasks against $O_t$, so oracle skills absent from the library count as unretrieved.
\emph{Bundle precision} is $|B_t \cap A_t| / |B_t|$ macro-averaged over $T_{+}$.
\emph{Dependency-pair co-recovery} is the fraction of induced dependency edges $(u,v)$ with $u, v \in A_t$ for which both endpoints appear in $B_t$.
Eleven of the 87 tasks have no oracle skill in the 200-skill library, so $|T_{+}| = 76$; availability-conditioned metrics use those 76 tasks and full-oracle recall uses all 87. Conditioning on availability keeps the 11 empty-oracle tasks out of those averages. They would otherwise contribute a vacuous perfect score for every arm.

\begin{table*}[!t]
\centering
\small
\caption{Complete matched-retrieval results, pooled over three independent graph builds. $R_{\mathrm{avail}}$ and $C$ use the 76 tasks with at least one in-library oracle skill; $R_{\mathrm{full}}$ uses all 87. Realized context size and bundle size show that the arms are matched in delivered payload, not only in nominal budget. Latency is pure retrieval time and excludes agent execution.}
\label{tab:appendix_matched_full}
\begin{ApdxTabFrame}
\apdxstyle
\begin{tabularx}{\linewidth}{@{}A{0.195\linewidth}C{0.068\linewidth}C{0.068\linewidth}C{0.058\linewidth}C{0.058\linewidth}C{0.068\linewidth}C{0.086\linewidth}C{0.064\linewidth}Y@{}}
\apdxband
\apdxhdr{Operation} & \apdxhdr{$R_{\mathrm{avail}}$} & \apdxhdr{$R_{\mathrm{full}}$} & \apdxhdr{$C$} & \apdxhdr{$D$} & \apdxhdr{Prec.} & \apdxhdr{Chars} & \apdxhdr{Skills} & \apdxhdr{p50/p95 (ms)} \\
\addlinespace[2.5pt]
Reverse-aware PPR & 0.575 & 0.502 & 0.421 & 0.654 & 0.280 & 11{,}999.8 & 4.86 & 10.89 / 12.30 \\
Forward-only PPR & 0.490 & 0.428 & 0.311 & 0.362 & 0.235 & 11{,}999.8 & 4.93 & 10.03 / 11.11 \\
No-Graph & 0.542 & 0.473 & 0.368 & 0.481 & 0.272 & 11{,}999.6 & 4.83 & \phantom{0}9.32 / 10.36 \\
One-Hop dependency & 0.542 & 0.473 & 0.368 & 0.481 & 0.276 & 11{,}991.0 & 4.79 & \phantom{0}9.63 / 10.39 \\
\end{tabularx}
\end{ApdxTabFrame}
\end{table*}

\begin{table*}[!t]
\centering
\small
\caption{Paired task-level bootstrap over the 76 availability-conditioned tasks, 5{,}000 resamples. The three graph replicates are averaged within each task before resampling. Reverse-aware PPR is ahead of every control on recall and completeness, whereas its precision advantage over the flat and one-hop controls is not distinguishable from zero.}
\label{tab:appendix_matched_bootstrap}
\begin{ApdxTabFrame}
\apdxstyle
\begin{tabularx}{\linewidth}{@{}A{0.19\linewidth}ZZZ@{}}
\apdxband
\apdxhdr{Comparison} & \apdxhdr{$\Delta R_{\mathrm{avail}}$ [95\% CI]} & \apdxhdr{$\Delta C$ [95\% CI]} & \apdxhdr{$\Delta$ Precision [95\% CI]} \\
\addlinespace[2.5pt]
Reverse $-$ Forward & $+0.085$ \, $[+0.044, +0.132]$ & $+0.110$ \, $[+0.048, +0.175]$ & $+0.045$ \, $[+0.021, +0.071]$ \\
Reverse $-$ No-Graph & $+0.033$ \, $[+0.004, +0.066]$ & $+0.053$ \, $[+0.009, +0.105]$ & $+0.008$ \, $[-0.009, +0.027]$ \\
Reverse $-$ One-Hop & $+0.033$ \, $[+0.003, +0.065]$ & $+0.053$ \, $[+0.004, +0.105]$ & $+0.004$ \, $[-0.015, +0.023]$ \\
\end{tabularx}
\end{ApdxTabFrame}
\end{table*}

\paragraph{Per-build results.}
Table~\ref{tab:appendix_matched_bootstrap} reports the paired intervals. The reverse-over-forward ordering holds on every individual graph rather than only in aggregate. Available-oracle recall for reverse-aware PPR versus forward-only PPR is $0.612$ versus $0.488$ on build A, $0.549$ versus $0.475$ on build B, and $0.563$ versus $0.506$ on build C. The spread across builds ($0.549$--$0.612$ for the reverse arm) is comparable in magnitude to the reverse-minus-No-Graph gap, so the graph-versus-flat margin is of the same order as construction variance, which Appendix~\ref{sec:appendix_graph_diagnostics} quantifies directly.

\begin{table}[htbp]
\centering
\small
\caption{Deterministic-core control. Restricting the graph to deterministically induced I/O dependency edges and removing all LLM-validated relations. Reverse traversal still improves completeness over forward propagation, while on recall the deterministic core matches rather than exceeds the matched flat retriever. Recomputed from the per-task records on the same 76-task availability-conditioned basis as Table~\ref{tab:appendix_matched_full}, so the two tables are directly comparable.}
\label{tab:appendix_deterministic_core}
\begin{ApdxTabFrame}
\apdxstyle
\begin{tabularx}{\linewidth}{@{}A{0.42\linewidth}A{0.14\linewidth}A{0.14\linewidth}Y@{}}
\apdxband
\apdxhdr{Operation} & \apdxhdr{$R_{\mathrm{avail}}$} & \apdxhdr{$C$} & \apdxhdr{$D$} \\
\addlinespace[2.5pt]
Reverse-aware PPR & 0.538 & 0.368 & 0.500 \\
Forward-only PPR & 0.512 & 0.316 & 0.342 \\
No-Graph & 0.542 & 0.368 & 0.474 \\
One-Hop dependency & 0.542 & 0.368 & 0.474 \\
\end{tabularx}
\end{ApdxTabFrame}
\end{table}

\paragraph{Deterministic core versus full graph.}
Table~\ref{tab:appendix_deterministic_core} repeats the comparison on a graph containing only the deterministically induced I/O dependency edges. Reverse traversal still improves exact completeness over forward propagation by $+0.053$ ($95\%$ CI $[+0.013, +0.105]$), and it still improves dependency-pair co-recovery from $0.342$ to $0.500$, but its recall advantage over forward propagation shrinks to $+0.026$ with an interval spanning zero ($[-0.007, +0.065]$), and against No-Graph it is $-0.004$ ($[-0.021, +0.015]$). Deterministic dependency structure alone therefore does not outperform flat retrieval. It is the LLM-validated relations that supply the coverage turning the deterministic backbone into a measurable retrieval advantage, which is what the full-graph results in Table~\ref{tab:appendix_matched_full} show.

\paragraph{Reading these numbers.}
Because the retrieval contract and the realized payload are held fixed, the differences in Table~\ref{tab:appendix_matched_full} are attributable to the graph operation alone, and they identify reverse-weighted traversal as the variant that recovers structurally coherent bundles. They are retrieval-level differences: no agent is run, so they describe which skills reach the prompt rather than what an agent then does with them, and the lexical seeding matches the deployed system of Table~\ref{results-table} only up to the seed stage. Dependency-pair co-recovery is likewise a structural measure, computed against a benchmark oracle that lists task-facing skills rather than a full prerequisite closure.

\section{Graph Quality, Cost, and Robustness}
\label{sec:appendix_graph_diagnostics}

\begin{table}[htbp]
\centering
\small
\caption{Graph statistics for the primary 200-skill graph and for a conservative re-gated replay of the same cached relation proposals. The replay trades coverage for precision: it removes 23 edges and raises the isolate count from 59 to 73, raising the isolate share from 29.5\% to 36.5\%, while reverse-PPR available-oracle recall slips from 0.575 to 0.567 on the 76-task basis. The replay reuses the node normalization and cached relation proposals of the primary build, so it isolates the effect of the acceptance gates.}
\label{tab:appendix_graph_stats}
\begin{ApdxTabFrame}
\apdxstyle
\begin{tabularx}{\linewidth}{@{}A{0.46\linewidth}ZZ@{}}
\apdxband
\apdxhdr{Statistic} & \apdxhdr{Primary} & \apdxhdr{Re-gated} \\
\addlinespace[2.5pt]
Nodes & 200 & 200 \\
Typed directed edges & 173 & 150 \\
\quad dependency & 110 & 88 \\
\quad workflow & 25 & 25 \\
\quad semantic & 30 & 29 \\
\quad alternative & 8 & 8 \\
Deterministic I/O edges & 73 & 66 \\
LLM-validated edges & 100 & 84 \\
Average total degree & 1.73 & 1.50 \\
Density & 0.004347 & 0.003769 \\
Isolated nodes & 59 & 73 \\
Giant component & 39/200 & 42/200 \\
Duplicate typed edges & 0 & 0 \\
Invalid endpoints & 0 & 0 \\
\end{tabularx}
\end{ApdxTabFrame}
\end{table}

\paragraph{Graph sparsity.}
The primary graph is sparse: average total degree is $1.73$ and $59$ of $200$ nodes carry no typed edge at all. For those isolated skills GoS falls back on its seeding and reranking path, which limits the margin it can gain over a budget-matched flat retriever. The sparsity follows from biasing construction toward precision---candidates are prefiltered and only high-confidence relations are accepted---and it bounds how much structural retrieval can contribute on any one library.

The re-gated replay in Table~\ref{tab:appendix_graph_stats} makes the trade-off explicit. Replaying the same cached relation proposals through the stricter gates that our error analysis produced removes 23 edges, most of them dependency edges, and lifts the isolate count from 59 to 73; available-oracle recall moves only from $0.575$ to $0.567$. Tightening acceptance therefore buys type and direction precision at a small cost in coverage, with retrieval close to unchanged in either setting. The paper reports the primary graph throughout. Because the replay reuses the node normalization and the cached LLM proposals of that build, it measures the effect of the gates themselves rather than of a fresh construction.

\begin{table}[htbp]
\centering
\small
\caption{Provenance-blinded 40-edge quality audit, stratified into eight edges per type/provenance stratum; cells give correct/scored counts. Direction (Dir.) is scored only on the 24 directional relations, dependency and workflow; semantic and alternative relations are non-directional and marked n/a. Labels were AI-drafted from the blinded packet and reviewed row by row by an author before the hidden key was opened.}
\label{tab:appendix_edge_audit}
\begin{ApdxTabFrame}
\apdxstyle
\begin{tabularx}{\linewidth}{@{}A{0.36\linewidth}ZZZ@{}}
\apdxband
\apdxhdr{Stratum} & \apdxhdr{Valid} & \apdxhdr{Type} & \apdxhdr{Dir.} \\
\addlinespace[2.5pt]
Overall (N=40) & 40/40 & 34/40 & 21/24 \\
\midrule
Deterministic dependency & 8/8 & 7/8 & 7/8 \\
LLM dependency & 8/8 & 6/8 & 7/8 \\
LLM workflow & 8/8 & 6/8 & 7/8 \\
LLM semantic & 8/8 & 7/8 & n/a \\
LLM alternative & 8/8 & 8/8 & n/a \\
\end{tabularx}
\end{ApdxTabFrame}
\end{table}

\paragraph{Edge quality: valid relations, imperfect type boundaries.}
Table~\ref{tab:appendix_edge_audit} reports the audit summarized in Section~\ref{sec:graph-diagnostics}, drawn from the conservatively re-gated graph of Table~\ref{tab:appendix_graph_stats} using a fresh SHA-256 stratified sample. Every sampled relation was judged operationally valid, but six of forty were better described by a different type: one deterministic dependency was an alternative; two LLM dependencies were a workflow and an alternative; one semantic edge was an alternative; and two workflow edges were a semantic and a dependency relation. All three direction errors are the same phenomenon---a relation proposed as directional whose best type is non-directional---rather than a genuinely reversed prerequisite. The practical implication is that type boundaries, not relation existence, are the weak point of LLM-validated construction. This matters because relation type determines the forward and reverse weights in Table~\ref{tab:appendix_relation_weights}.

An earlier 40-edge packet on the primary graph gave validity, type, and direction of $39/40$, $36/40$, and $23/24$. That packet was used to develop the acceptance gates and then to measure them, so it functions as development error analysis; the held-out audit above supplies the graph-quality estimate. Its diagnostic value was concrete: it exposed generic-container matches, tooling-only dependencies, weak cross-domain matches, result-state words promoted to artifacts, and generic automation-wrapper semantics, each of which became a regression test and a construction gate.

Both audits measure precision: they estimate what fraction of the asserted edges are correct, and are silent on how many true relations construction missed. Measuring that recall would require a human-annotated prerequisite graph for SkillsBench, which does not exist.

\begin{table*}[!t]
\centering
\small
\caption{Construction stability. Relation-only rebuilds share one frozen normalized node manifest, whereas end-to-end rebuilds also repeat LLM semantic completion of missing node fields. Freezing the manifest makes the deterministic candidate set exactly reproducible and roughly doubles typed-edge agreement, which is why we version normalized manifests rather than regenerating them.}
\label{tab:appendix_construction_stability}
\begin{ApdxTabFrame}
\apdxstyle
\begin{tabularx}{\linewidth}{@{}A{0.28\linewidth}A{0.30\linewidth}Y@{}}
\apdxband
\apdxhdr{Quantity} & \apdxhdr{Relation-only (frozen manifest)} & \apdxhdr{End-to-end (repeated normalization)} \\
\addlinespace[2.5pt]
Edges per build & 173 / 158 / 159 & 196 / 191 / 175 \\
Mean typed-directed edge Jaccard & 0.607 & 0.297 \\
Pairwise typed-directed Jaccard & 0.607 / 0.596 / 0.617 & 0.299 / 0.284 / 0.307 \\
Deterministic candidate Jaccard & 1.000 / 1.000 / 1.000 & 0.358 / 0.304 / 0.404 \\
LLM-validated edge Jaccard & 0.355--0.375 & --- \\
\end{tabularx}
\end{ApdxTabFrame}
\end{table*}

\paragraph{Where construction variance comes from.}
Table~\ref{tab:appendix_construction_stability} separates the two sources of variance. Three relation builds over one frozen normalized manifest produce 173/158/159 edges with mean typed-directed Jaccard $0.607$; all three ran with MiniMax M2.7 at temperature $0.0$, with complete-response caching disabled and an identical construction fingerprint. The deterministic candidate set is bit-identical across all three (Jaccard $1.000$), so all of the disagreement lives in LLM relation validation, whose pairwise edge Jaccard is $0.355$--$0.375$. The provider exposes no reproducible sampling seed, so stability is measured here by empirical repeats at temperature $0.0$.

Repeating the entire pipeline is markedly less stable. Node names and parser-owned fields are perfectly reproducible: description and script entrypoints match on all $200/200$ nodes. LLM-completed fields are not. Pairwise set Jaccard is $0.328$ for inputs, $0.349$ for outputs, $0.524$ for domain tags, $0.559$ for tooling, and $0.265$ for example tasks, and only $30$/$34$/$32$/$48$/$24$ of $200$ nodes are exactly identical across all three runs.

Deterministic dependency induction reads exactly those inferred I/O fields, so node-level variance propagates into the graph and the deterministic candidate set is no longer stable (Jaccard $0.304$--$0.404$). The operational conclusion is that semantic completion, not relation validation, is the dominant source of end-to-end irreproducibility, and that normalized manifests should be persisted and versioned rather than regenerated on each rebuild.

\begin{table}[htbp]
\centering
\small
\caption{Dependency threshold sensitivity on the primary graph, with retrieval scored on the same 76-task availability-conditioned basis as Table~\ref{tab:appendix_matched_full}. Edge counts are re-induced candidates, so the $\zeta{=}0.6$ row gives 75 where the persisted graph keeps 73: two candidates are shadowed by an LLM-validated edge carrying the same typed dependency key. Lowering $\zeta$ admits nearly four times as many deterministic edges but leaves recall essentially unchanged, while raising $\zeta$ costs exact completeness. $\zeta{=}0.6$ is held fixed by design across all experiments.}
\label{tab:appendix_zeta}
\begin{ApdxTabFrame}
\apdxstyle
\begin{tabularx}{\linewidth}{@{}A{0.10\linewidth}A{0.16\linewidth}A{0.22\linewidth}Y@{}}
\apdxband
\apdxhdr{$\zeta$} & \apdxhdr{Edges} & \apdxhdr{Recall} & \apdxhdr{Complete} \\
\addlinespace[2.5pt]
0.4 & 172 & 0.571 & 0.421 \\
0.6 & \phantom{0}75 & 0.575 & 0.408 \\
0.8 & \phantom{0}46 & 0.574 & 0.395 \\
\end{tabularx}
\end{ApdxTabFrame}
\end{table}

\begin{table*}[!t]
\centering
\small
\caption{Offline construction telemetry. $Nk$ is an upper bound on submitted candidate \emph{pairs}, not on API calls: candidates for one focus skill are batched into a single structured validator request, so requests are at most $N$ before retries. The two builds differ in concurrency and in normalized-metadata completeness, so the columns report operational cost at two library sizes rather than a controlled scaling curve.}
\label{tab:appendix_construction_cost}
\begin{ApdxTabFrame}
\apdxstyle
\begin{tabularx}{\linewidth}{@{}A{0.38\linewidth}A{0.22\linewidth}Y@{}}
\apdxband
\apdxhdr{Quantity} & \apdxhdr{200 skills} & \apdxhdr{2{,}000 skills} \\
\addlinespace[2.5pt]
$Nk$ candidate upper bound & 1{,}600 & 16{,}000 \\
Submitted candidate pairs & 1{,}287 & 7{,}060 \\
Batched validator requests & 192 & 1{,}050 \\
Total provider calls & 195 & 1{,}091 \\
Relation proposals returned & 172 & 2{,}662 \\
\quad accepted / rejected & 118 / 54 & 2{,}504 / 158 \\
\quad accepted duplicates dropped & 18 & 16 \\
Input tokens & 544{,}112 & 4{,}999{,}552 \\
Output tokens & 173{,}763 & 1{,}090{,}845 \\
Cached-input tokens & 245{,}993 & 95{,}254 \\
Reasoning tokens & 153{,}702 & 727{,}738 \\
Relink concurrency & 2 & 6 \\
Wall time & 37.1 min & 85.2 min \\
Provider cost & \$0.2855 & \$2.3596 \\
Nodes complete in I/O and semantics & 200/200 & 477/2{,}000 (23.8\%) \\
\end{tabularx}
\end{ApdxTabFrame}
\end{table*}

\paragraph{Construction accounting.}
Table~\ref{tab:appendix_zeta} reports the threshold sweep and Table~\ref{tab:appendix_construction_cost} the construction telemetry. The complexity statement in Section~\ref{sec:methodology} should be read as bounding submitted candidate pairs. In the 200-skill build that bound was $Nk = 1{,}600$, and prefiltering brought the pairs actually submitted down to $1{,}287$, which were batched into $192$ validator requests. That request count falls below $N$ because a focus skill with no surviving candidate issues no request at all. Of the $172$ relation proposals the validator returned, deterministic post-validation gates accepted $118$ and rejected $54$; deduplication then removed a further $18$ of the accepted proposals, leaving exactly the $100$ LLM-validated edges of Table~\ref{tab:appendix_graph_stats}. Construction is a one-time offline cost of well under a dollar at this scale, and roughly \$2.36 at 2{,}000 skills.

The 2{,}000-skill column differs from the 200-skill build in two respects: it ran at a different concurrency setting, so its wall time is not comparable, and only $477$ of its $2{,}000$ nodes are complete in both I/O and semantic metadata, against $200/200$ for the curated 200-skill library. The column therefore reports construction cost at that scale over a corpus of mixed metadata quality, with $N$ and corpus quality varying together.

\paragraph{Candidate prefiltering.}
Candidate preparation uses an exact evidence-token postings index rather than scoring all pairs. On a fixed-hash 100-node sample of the 2{,}000-skill library, the index reduced the scored pairs from $199{,}900$ to $37{,}618$ ($-81.2\%$) and cut CPU time from $33.22$\,s to $12.41$\,s, a $2.68\times$ speedup that already includes a $0.07$\,s index build. For every sampled node, the top-32 candidate ranking is reproduced exactly.

\begin{table}[htbp]
\centering
\small
\caption{Metadata corruption stress test on the deterministic graph core, under fixed-hash selection of the corrupted subset. Description quality dominates: removing half of all I/O fields costs 0.024 recall, whereas removing or genericizing half of the descriptions costs roughly 0.15. This harness scores all 87 queries and uses a 9{,}000-character bundle budget, so its absolute values are not comparable with Table~\ref{tab:appendix_matched_full}; the deltas within the table are the quantity of interest.}
\label{tab:appendix_metadata_stress}
\begin{ApdxTabFrame}
\apdxstyle
\begin{tabularx}{\linewidth}{@{}A{0.26\linewidth}A{0.10\linewidth}A{0.15\linewidth}A{0.15\linewidth}Y@{}}
\apdxband
\apdxhdr{Corruption} & \apdxhdr{Frac.} & \apdxhdr{Edges} & \apdxhdr{$R_{\mathrm{avail}}$} & \apdxhdr{$\Delta$} \\
\addlinespace[2.5pt]
None (baseline) & --- & 314 & 0.589 & --- \\
\midrule
Drop I/O fields & 10\% & 246 & 0.591 & $+0.002$ \\
 & 25\% & 164 & 0.561 & $-0.028$ \\
 & 50\% & \phantom{0}71 & 0.565 & $-0.024$ \\
\midrule
Drop descriptions & 10\% & 314 & 0.551 & $-0.038$ \\
 & 25\% & 314 & 0.485 & $-0.104$ \\
 & 50\% & 314 & 0.443 & $-0.146$ \\
\midrule
Genericize descriptions & 10\% & 314 & 0.551 & $-0.038$ \\
 & 25\% & 314 & 0.485 & $-0.104$ \\
 & 50\% & 314 & 0.447 & $-0.142$ \\
\end{tabularx}
\end{ApdxTabFrame}
\end{table}

\paragraph{Metadata quality dominates I/O structure.}
Table~\ref{tab:appendix_metadata_stress} stresses the deterministic graph core by corrupting normalized node metadata. Dropping I/O fields is destructive to the graph---half the deterministic edges disappear by the $25\%$ condition and $77\%$ of them by the $50\%$ condition---yet retrieval degrades only mildly, because seeding and reranking still find the right entry points. Corrupting descriptions leaves the deterministic edge count untouched at $314$ but is roughly six times more damaging to recall, because descriptions carry the query-to-seed signal as well as the semantic-completion signal. GoS therefore inherits documentation quality more than schema completeness, and a library with rich descriptions but sparse I/O frontmatter is a better substrate than the converse. The deltas are offline retrieval effects; how they translate into agent reward is not measured here.

\paragraph{Corpus coverage and ALFWorld provenance.}
The 87 SkillsBench tasks reference 195 unique oracle skill directories. Each evaluated collection---200, 500, 1{,}000, and 2{,}000 skills---contains 157 exact matches plus four conservative punctuation aliases (for example \skillcode{dialogue-graph} against \skillcode{dialogue_graph}). A further 34 are absent from every collection. Consequently $75/87$ tasks ($86.2\%$) have their entire oracle available, and 11 tasks have no oracle skill in the 200-skill library at all. The latter is the origin of the 76-task availability-conditioned denominator used throughout Appendix~\ref{sec:appendix_matched_retrieval}. All retrieval arms draw from the same collection, so the gap is shared rather than differential. It does, however, bound the absolute recall values below their theoretical maximum, and it makes them specific to this corpus.

For ALFWorld, the 200-skill collection contains no ALFWorld skills, while the 500, 1{,}000, and 2{,}000 collections each contain 37. Raw ALFWorld packages carry descriptions, but they do not uniformly provide formal I/O or domain frontmatter. Graph construction therefore relies on conservative semantic completion with field-level provenance rather than on assuming the fields were observed. In the 1{,}000-node extraction, 36, 35, and 34 of the 37 packages have inputs, outputs, and domain tags respectively. Each inferred field carries an inferred provenance flag, which keeps it distinct from parser-owned values downstream. This is the main reason the ALFWorld blocks in Table~\ref{results-table} depend more on seeding quality than on deep dependency structure.

\section{Error Analysis}

\paragraph{Error Taxonomy.}
We distinguish retrieval-side errors from downstream execution failures, since these correspond to different limits of the overall system. A retrieval method can fail because it never surfaces the correct skill, because it retrieves an incomplete bundle that omits critical prerequisites, or because it produces a broadly adequate bundle that is subsequently misused by the downstream agent. Treating these failure modes separately is important for attributing gains correctly: GoS is designed to improve retrieval completeness, but it cannot by itself eliminate planning or execution failures once a bundle has already been provided. Across our trajectories, several of the most informative failures are not simple retrieval misses, but long-horizon search failures in which the correct general tool family is present and the agent still does not converge to a verifier-passing implementation.

\begin{table}[t]
\centering
\small
\caption{Primary error modes in GoS-style retrieval experiments. The table separates retrieval failures, downstream execution failures, and infrastructure issues.}
\label{tab:appendix_error_modes}
\begin{ApdxTabFrame}
\apdxstyle
\begin{tabularx}{\linewidth}{@{}A{0.24\linewidth}A{0.38\linewidth}Y@{}}
\apdxband
\apdxhdr{Error type} & \apdxhdr{Typical symptom} & \makecell[l]{\textcolor{white}{\textbf{Whether helps}}} \\
\addlinespace[2.5pt]
\midrule
Retrieval miss & The correct skill exists in the library but is not retrieved, forcing a from-scratch path. & Yes; better seeding and graph completion reduce these failures. \\
Partial retrieval & A relevant skill is retrieved, but supporting parser, setup, or constraint prerequisites are absent. & Yes; this is the main failure mode GoS is designed to prevent. \\
Good retrieval, bad execution & The retrieved bundle is adequate, but the agent drifts, over-engineers, or mismatches the verifier. & Only indirectly; this is a planning or backbone issue. \\
Infrastructure failure & Build, environment, or startup failures prevent execution. & No; these are excluded from quality comparisons. \\
\end{tabularx}
\end{ApdxTabFrame}
\end{table}

\paragraph{Retrieval Misses.}
Table~\ref{tab:appendix_error_modes} defines the four categories. A retrieval miss occurs when the correct skill exists in the repository but is not surfaced at all. In this regime, the agent is forced onto a generic from-scratch path, so any downstream failure should be attributed primarily to the retriever rather than to execution drift. Misses typically arise when the query language does not overlap strongly with the skill description, when the task is phrased around a downstream objective but the critical skill is an upstream parser or setup utility, or when semantic retrieval overweights topical similarity relative to executable relevance. A retrieval miss has a characteristic consequence: once the relevant analysis skill is absent, the task degrades into from-scratch implementation and fails the verifier.

\paragraph{Partial Retrieval.}
Partial retrieval is more subtle and often more important. Here, the retrieved bundle contains an obviously relevant high-level skill, but omits one or more prerequisite helpers needed for successful completion. In our setting, these missing items are often parsers, format converters, preprocessing utilities, setup routines, or constraint-carrying reference skills. This is precisely the regime in which graph-aware retrieval is intended to help: once a high-level skill is matched as a seed, reverse-aware propagation can recover supporting skills that are weak semantic matches to the raw query but strong structural neighbors in the skill graph. \texttt{earthquake-phase-association} illustrates the boundary of this idea. In the stronger all-skills trajectory, the agent assembled a coherent seismic stack including \texttt{gamma-phase-associator}, \texttt{obspy-data-api}, \texttt{seisbench-model-api}, and \texttt{seismic-picker-selection}, and the task passed. In the corresponding GoS case, the graph retrieval did bring in a partially relevant seismic bundle, but the resulting context was still less complete, and the task failed with reward $0.0$. This suggests that structural retrieval helps only when the recovered neighborhood is sufficiently complete to support the downstream pipeline, not merely when one or two domain-relevant skills are present.

\paragraph{Good Retrieval, Bad Execution.}
Some failures occur even when the retrieved bundle is broadly adequate. In these cases, the agent may still over-generalize the task, ignore a retrieved authoritative interface, implement unnecessary functionality, or fail to align with the verifier. These episodes matter because they bound what can be credited to retrieval alone. They also motivate the conservative agent-facing instructions used in our environments, which emphasize verifier-minimal solutions, direct use of returned local source paths, and avoidance of unnecessary branching. \texttt{energy-market-pricing} is a representative example: both all-skills and GoS had access to the relevant economic-dispatch / power-flow skill family and both eventually passed, but the all-skills trajectory required substantially more agent time before converging. This is not a retrieval miss; it is a difference in how efficiently a broadly adequate bundle is converted into a verifier-passing plan. Conversely, \texttt{adaptive-cruise-control} shows the opposite failure mode: the retrieved bundle included clearly relevant control skills such as \texttt{pid-controller}, \texttt{mpc-horizon-tuning}, \texttt{vehicle-dynamics}, and \texttt{simulation-metrics}, yet the run still finished with reward $0.0$. In that case the failure is better described as long-horizon execution drift or verifier mismatch rather than poor retrieval.

A related example is \texttt{dialogue-parser}. Multiple conditions had access to relevant task structure, yet only the strongest GoS trajectory converted that context into a full pass, while other conditions remained at partial reward. This again indicates that the dominant bottleneck was not the absence of any relevant skill at all, but how effectively the agent translated the available skill context into the exact output expected by the verifier.

\paragraph{Infrastructure Failures.}
Finally, a subset of observed failures are not retrieval failures at all, but infrastructure failures such as environment build issues, setup crashes, or startup timeouts before a substantive episode begins. Representative examples include Docker / BuildKit failures such as \texttt{layer does not exist} on \texttt{dapt-intrusion-detection}, missing compiler toolchains for \texttt{obspy}-dependent tasks, and logging failures that leave some trials with incomplete session traces or null token fields.

These cases are methodologically important but conceptually distinct. They matter operationally because they require reruns and infrastructure fixes, and they should be tracked for experiment hygiene and rerun logic. They are not, however, evidence about the retrieval quality of GoS, vector retrieval, or all-skills, and they should not be interpreted as evidence against the quality of the underlying model. We therefore treat them as hygiene issues and exclude them from method-quality interpretation.

\section{Qualitative Analysis}
\label{sec:appendix_qualitative}

\paragraph{Section Framing.}
We next examine a set of trajectory-grounded qualitative cases and compare the concrete skill evidence that actually entered the agent's working context in each condition. Table~\ref{tab:appendix_qualitative_retrieval} reports the skills that materially shaped each run: for GoS and Vector Skills, these are the skills surfaced by the retrieval call and then used downstream, while for \textit{Vanilla Skills} they are the skills the agent explicitly opened from the mounted library. This grounds comparisons in executed trajectories rather than hypothetical retrieval.

Across the cases below, we focus on a single question: does the method expose a compact, execution-ready bundle early enough to change the trajectory? The main qualitative difference is often not whether a relevant skill exists somewhere in the repository, but whether the agent receives the right subset in a form that can be operationalized under the task budget.

\begingroup
\newcommand{\qskill}[1]{\texttt{#1}}
\newcommand{\qcell}[1]{\parbox[t]{\linewidth}{\raggedright #1}}
\begin{table*}[t]
\centering
\scriptsize
\caption{Trajectory-grounded skill evidence from executed qualitative cases. \textsc{Useful} denotes skills operationalized downstream; \textsc{Noisy} denotes items that were tangential or not used.}
\label{tab:appendix_qualitative_retrieval}
\begin{ApdxTabFrame}[width=\textwidth]
\begin{tabularx}{\textwidth}{>{\raggedright\arraybackslash}p{0.145\textwidth}>{\raggedright\arraybackslash}X>{\raggedright\arraybackslash}X>{\raggedright\arraybackslash}X}
\apdxband
\apdxhdr{Task} & \apdxhdr{GoS Bundle} & \apdxhdr{Vanilla Bundle} & \apdxhdr{Vector Bundle} \\
\addlinespace[2.5pt]
\qcell{\texttt{pedestrian-}\\\texttt{traffic-}\\\texttt{counting}} &
\qcell{\textsc{Useful} \qskill{gemini-count-in-video}; \qskill{multimodal-fusion}; \qskill{openai-vision}; \qskill{video-frame-extraction}\\\textsc{Noisy} \qskill{threat-detection}} &
\qcell{\textsc{Useful} \qskill{gemini-count-in-video}; \qskill{object\_counter}; \qskill{openai-vision}; \qskill{video-frame-extraction}\\\textsc{Noisy} \qskill{alfworld-heat-object-}\\\qskill{with-appliance}; \qskill{alfworld-object-locator}; broader noisy context} &
\qcell{\textsc{Useful} none\\\textsc{Noisy} \qskill{google-classroom-automation}; \qskill{rdkit}; \qskill{salesforce-service-cloud-}\\\qskill{automation}; \qskill{segmetrics-automation}} \\

\qcell{\texttt{flood-risk-}\\\texttt{analysis}} &
\qcell{\textsc{Useful} \qskill{flood-detection}; \qskill{nws-flood-thresholds}; \qskill{usgs-data-download}\\\textsc{Noisy} \qskill{time\_series\_anomaly\_detection}; \qskill{-21risk-automation}} &
\qcell{\textsc{Useful} \qskill{flood-detection}; \qskill{nws-flood-thresholds}; \qskill{usgs-data-download}} &
\qcell{\textsc{Useful} none\\\textsc{Noisy} \qskill{leverly-automation}; \qskill{scienceworld-room-navigator}; \qskill{text-to-speech}; broader noisy context} \\

\qcell{\texttt{travel-}\\\texttt{planning}} &
\qcell{\textsc{Useful} \qskill{search-accommodations}; \qskill{search-attractions}; \qskill{search-cities}; \qskill{search-driving-distance}; \qskill{search-restaurants}\\\textsc{Noisy} \qskill{search-flights}; \qskill{fjsp-baseline-repair-}\\\qskill{with-downtime-and-policy}} &
\qcell{\textsc{Useful} \qskill{search-accommodations}; \qskill{search-attractions}; \qskill{search-cities}; \qskill{search-restaurants}} &
\qcell{\textsc{Useful} \qskill{search-accommodations}; \qskill{search-attractions}; \qskill{search-cities}; \qskill{search-driving-distance}; \qskill{search-restaurants}\\\textsc{Noisy} additional noisy items} \\

\qcell{\texttt{dapt-}\\\texttt{intrusion-}\\\texttt{detection}} &
\qcell{\textsc{Useful} \qskill{pcap-analysis}; \qskill{pcap-triage-tshark}\\\textsc{Noisy} \qskill{dc-power-flow}; \qskill{power-flow-data}; \qskill{-21risk-automation}} &
\qcell{\textsc{Useful} no clearly reused core skill} &
\qcell{\textsc{Useful} none\\\textsc{Noisy} \qskill{codacy-automation}; \qskill{jakarta-namespace}; \qskill{rootly-automation}; broader noisy context} \\

\qcell{\texttt{dialogue-}\\\texttt{parser}} &
\qcell{\textsc{Useful} \qskill{dialogue\_graph}; \qskill{webshop-query-parser}\\\textsc{Noisy} \qskill{browser-testing}; \qskill{obj-exporter}; \qskill{alfworld-goal-interpreter}} &
\qcell{\textsc{Useful} \qskill{dialogue\_graph}} &
\qcell{\textsc{Useful} none\\\textsc{Noisy} \qskill{docnify-automation}; \qskill{scienceworld-task-focuser}; \qskill{temporal-python-testing}; broader noisy context} \\

\qcell{\texttt{earthquake-}\\\texttt{phase-}\\\texttt{association}} &
\qcell{\textsc{Useful} \qskill{gamma-phase-associator}; \qskill{seisbench-model-api}; \qskill{seismic-picker-selection}\\\textsc{Noisy} \qskill{flood-detection}; \qskill{-21risk-automation}} &
\qcell{\textsc{Useful} \qskill{gamma-phase-associator}; \qskill{obspy-data-api}; \qskill{obspy-datacenter-client}; \qskill{seisbench-model-api}; \qskill{seismic-picker-selection}\\\textsc{Noisy} \qskill{gamma-automation}; \qskill{seismic-automation}} &
\qcell{\textsc{Useful} none\\\textsc{Noisy} \qskill{fixer-automation}; \qskill{maven-build-lifecycle}; \qskill{segmetrics-automation}; broader noisy context} \\

\qcell{\texttt{energy-}\\\texttt{market-}\\\texttt{pricing}} &
\qcell{\textsc{Useful} \qskill{dc-power-flow}; \qskill{power-flow-data}; \qskill{locational-marginal-}\\\qskill{prices}; \qskill{casadi-ipopt-nlp}\\\textsc{Noisy} \qskill{-21risk-automation}} &
\qcell{\textsc{Useful} \qskill{dc-power-flow}; \qskill{economic-dispatch}} &
\qcell{\textsc{Useful} \qskill{power-flow-data}\\\textsc{Noisy} \qskill{aryn-automation}; \qskill{moxie-automation}; \qskill{mural-automation}; broader noisy context} \\

\qcell{\texttt{3d-scan-}\\\texttt{calc}} &
\qcell{\textsc{Useful} \qskill{mesh-analysis}; \qskill{dyn-object-masks}\\\textsc{Noisy} \qskill{scienceworld-circuit-builder}; \qskill{scienceworld-circuit-}\\\qskill{connector}; \qskill{scienceworld-conductivity-}\\\qskill{tester}} &
\qcell{\textsc{Useful} \qskill{mesh-analysis}\\\textsc{Noisy} broader noisy context} &
\qcell{\textsc{Useful} \qskill{mesh-analysis}; \qskill{obj-exporter}; \qskill{pymatgen}; \qskill{threejs}\\\textsc{Noisy} \qskill{reflow-profile-compliance-}\\\qskill{toolkit}} \\

\qcell{\texttt{adaptive-}\\\texttt{cruise-}\\\texttt{control}} &
\qcell{\textsc{Useful} \qskill{imc-tuning-rules}; \qskill{pid-controller}; \qskill{safety-interlocks}; \qskill{vehicle-dynamics}\\\textsc{Noisy} \qskill{-21risk-automation}} &
\qcell{\textsc{Useful} no clearly reused core skill} &
\qcell{\textsc{Useful} \qskill{pid-controller}; \qskill{mpc-horizon-tuning}; \qskill{integral-action-design}; \qskill{simulation-metrics}; \qskill{vehicle-dynamics}} \\

\qcell{\texttt{econ-}\\\texttt{detrending-}\\\texttt{correlation}} &
\qcell{\textsc{Useful} \qskill{timeseries-detrending}\\\textsc{Noisy} \qskill{artifacts-builder}; \qskill{dyn-object-masks}; \qskill{mesh-analysis}; \qskill{-21risk-automation}} &
\qcell{\textsc{Useful} no clearly reused core skill} &
\qcell{\textsc{Useful} none\\\textsc{Noisy} \qskill{breezy-hr-automation}; \qskill{scienceworld-object-}\\\qskill{classifier}; \qskill{webshop-purchase-}\\\qskill{initiator}; broader noisy context} \\
\end{tabularx}
\end{ApdxTabFrame}
\end{table*}
\endgroup

\paragraph{Case Study 1: Pedestrian Traffic Counting.} The clearest intermediate case in our trajectories is \texttt{pedestrian-traffic-counting}. The task requires frame extraction, reliable pedestrian counting, and structured output generation. GoS surfaced a compact visual pipeline centered on \texttt{gemini-count-in-video}, \texttt{video-frame-extraction}, and \texttt{openai-vision}, and achieved the strongest outcome among the three conditions ($0.417$). The \textit{Vanilla Skills} baseline did eventually open relevant helpers, including \texttt{gemini-count-in-video}, \texttt{video-frame-extraction}, and \texttt{object\_counter}, but reached only a partial score ($0.267$). The \textit{Vector Skills} run performed worst ($0.041$): although it issued the retrieval call, the retrieved context was not converted into a workable plan. This example is useful because it is not a pure pass/fail contrast. \textit{Vanilla Skills} does locate relevant functionality, but GoS exposes a smaller, more coherent bundle that is easier to operationalize under the task budget.

\paragraph{Case Study 2: Flood Risk Analysis.} The \texttt{flood-risk-analysis} task illustrates a different regime: both GoS and \textit{Vanilla Skills} succeed, but GoS exposes the required chain with much less search friction. In this task the correct workflow is not generic time-series analysis; it is specifically the combination of \texttt{usgs-data-download} for measurements, \texttt{nws-flood-thresholds} for stage cutoffs, and \texttt{flood-detection} for aggregation and comparison. GoS surfaced exactly this bundle and passed with reward $1.0$. The \textit{Vanilla Skills} baseline also passed with reward $1.0$, but only after the agent explicitly searched through the large mounted library and opened the same family of skills. \textit{Vector Skills}, by contrast, issued the retrieval command but never translated retrieval into a usable flood-analysis bundle, and the run failed with reward $0.0$. Despite the absence of a reward gap, this case shows that when the right execution chain exists in the repository, GoS helps by making it explicit earlier in the trajectory.

\paragraph{Case Study 3: Travel Planning.} The \texttt{travel-planning} task is informative precisely because all three conditions surfaced clearly relevant travel skills. In the GoS run, the retrieved context centered on \texttt{search-cities}, \texttt{search-accommodations}, \texttt{search-attractions}, \texttt{search-driving-distance}, and \texttt{search-restaurants}, which is very close to the intended tool chain for the task. The \textit{Vanilla Skills} baseline likewise opened essentially the same family of skills after searching through the library. \textit{Vector Skills} also surfaced and used this same cluster of \texttt{search-*} skills, and that run passed the verifier with reward $1.0$. This example sharpens the qualitative claim of the paper. The advantage of GoS is not that flat semantic retrieval can never recover the correct skill family; rather, it is that GoS more reliably exposes a compact and coherent bundle early in the episode. When \textit{Vector Skills} does succeed, as it does here, its behavior becomes qualitatively much closer to GoS than to a clean retrieval miss.

\paragraph{Case Study 4: Network Intrusion Detection.} A clean GoS-positive example is \texttt{dapt-intrusion-detection}. In this case, GoS surfaced \texttt{pcap-analysis} together with adjacent analysis helpers such as \texttt{pcap-triage-tshark} and \texttt{threat-detection}, and the task passed. By contrast, the corresponding vector condition retrieved unrelated automation-oriented skills rather than a usable PCAP analysis bundle, while the all-skills condition still failed despite full library access. This case complements the retrieval-miss pattern in the Error Analysis section: once retrieval surfaces the right analysis bundle, the task becomes a reuse problem rather than a from-scratch reverse-engineering problem.

\paragraph{Case Study 5: Dialogue Parsing.} The \texttt{dialogue-parser} examples exhibit a strong gradient across methods. GoS converted the task into a full pass while exposing a compact bundle centered on \texttt{dialogue\_graph}, together with structural helpers such as \texttt{obj-exporter}, \texttt{browser-testing}, and parser-oriented support. \textit{Vanilla Skills} eventually improved once it explicitly opened \texttt{dialogue\_graph}, and \textit{Vector Skills} also reached a substantial partial score, but neither condition showed the same level of structured completeness as the strongest GoS trajectory. This case illustrates a pattern that appears repeatedly: once the right latent-representation skill is surfaced early, the rest of the pipeline becomes much easier for the agent to operationalize.

\paragraph{Case Study 6: Earthquake Phase Association.} \texttt{earthquake-phase-association} is a useful negative case for GoS because it shows that structural retrieval does not help automatically when the recovered neighborhood is still incomplete. In the strongest all-skills trajectory, the agent assembled a seismic processing stack including \texttt{gamma-phase-associator}, \texttt{obspy-data-api}, \texttt{obspy-datacenter-client}, \texttt{seisbench-model-api}, and \texttt{seismic-picker-selection}, and the task passed. The corresponding GoS case surfaced only a weaker subset, centered on \texttt{gamma-phase-associator}, \texttt{seisbench-model-api}, and \texttt{seismic-picker-selection}, with an irrelevant distraction skill mixed into the bundle, and the task failed. This is exactly the kind of case that is easy to miss if one looks only at whether some domain-relevant skill was retrieved. The qualitative difference is that the all-skills run assembled a more complete stack, while GoS remained one step short of the required pipeline.

\paragraph{Case Study 7: Energy Market Pricing.} A final useful case is \texttt{energy-market-pricing}, where all-skills and GoS both passed but with very different trajectory quality. The all-skills condition explicitly used \texttt{dc-power-flow} and \texttt{economic-dispatch}, while GoS surfaced a broader but still coherent optimization bundle including \texttt{dc-power-flow}, \texttt{power-flow-data}, \texttt{locational-marginal-prices}, and \texttt{casadi-ipopt-nlp}. Both runs eventually passed the verifier, but the trajectory quality differed sharply: GoS converted the retrieved bundle into a solution with substantially less agent-side search. Here, the main value of GoS is not a higher reward, but a shorter path from retrieval to execution.

\paragraph{Case Study 8: 3D Scan Calculation.} The \texttt{3d-scan-calc} task serves as a useful control because all three conditions can succeed when they recover the same latent geometry bottleneck. GoS exposed \texttt{mesh-analysis} together with adjacent geometric helpers, directly matching the preprocessing structure of the task. \textit{Vanilla Skills} also reached a passing solution once the agent opened \texttt{mesh-analysis}, but the surrounding library context was notably noisier. \textit{Vector Skills} likewise passed when it surfaced \texttt{mesh-analysis} together with geometry-oriented companions such as \texttt{obj-exporter}, \texttt{pymatgen}, and \texttt{threejs}. The qualitative lesson is therefore not that GoS is uniquely capable of solving the task; rather, when all methods recover a geometry-centered bundle, all can succeed, and the remaining difference is how directly that bundle is exposed.

\paragraph{Case Study 9: Adaptive Cruise Control.} \texttt{adaptive-cruise-control} is a useful failure case because all three conditions surfaced highly plausible control-related skills and still failed. GoS exposed \texttt{imc-tuning-rules}, \texttt{pid-controller}, \texttt{safety-interlocks}, and \texttt{vehicle-dynamics}, while \textit{Vector Skills} surfaced an even more explicit control bundle including \texttt{pid-controller}, \texttt{mpc-horizon-tuning}, \texttt{integral-action-design}, \texttt{simulation-metrics}, and \texttt{vehicle-dynamics}. The \textit{Vanilla Skills} condition also had access to a broad control-oriented context, yet none of the three settings converged to a passing solution. This case is important precisely because it is not a retrieval miss. It shows that once the task requires a long control-design and verifier-alignment chain, even a qualitatively good skill bundle may not be enough; the dominant bottleneck shifts from retrieval to execution discipline.

\paragraph{Case Study 10: Economic Detrending and Correlation.} The \texttt{econ-detrending-correlation} task offers a complementary success case. GoS surfaced \texttt{timeseries-detrending} and converted the task into a full pass, while the all-skills condition failed to assemble a comparably coherent detrending-centered bundle. \textit{Vector Skills} also reached a full pass, but with a noisier retrieved context whose skills were only weakly connected to the intended econometric workflow. This case is useful in two ways. First, it shows another task where surfacing the right latent preprocessing step, here detrending rather than raw correlation, materially changes the result. Second, it reinforces the lesson that vector retrieval can succeed, but its successful runs do not always arise from a bundle as semantically crisp or structurally interpretable as that of GoS.

\paragraph{Takeaway.} Across all ten qualitative cases in this appendix, the main pattern is not simply that GoS retrieves skills with better topical overlap. Rather, GoS more often exposes a bundle that is already close to the executable decomposition of the task. The \texttt{pedestrian-traffic-counting} example shows a genuine middle case in which \textit{Vanilla Skills} finds relevant tools but still underperforms the tighter GoS bundle. The \texttt{flood-risk-analysis} example shows that when the correct chain is available to multiple methods, GoS mainly reduces search friction and makes the intended execution path explicit earlier. The \texttt{travel-planning}, \texttt{3d-scan-calc}, and \texttt{econ-detrending-correlation} examples show that \textit{Vector Skills} can also succeed when it recovers the right family of skills, but these successes are most convincing when the retrieved bundle becomes qualitatively similar to what GoS surfaces directly. The \texttt{dialogue-parser} and \texttt{dapt-intrusion-detection} cases show how GoS can convert that structural advantage into clearer downstream wins. 

By contrast, \texttt{earthquake-phase-association} shows a real boundary condition in which GoS still falls short of an execution-complete bundle, while \texttt{energy-market-pricing} and \texttt{adaptive-cruise-control} show that even with broadly adequate retrieval, trajectory efficiency and verifier alignment remain separate bottlenecks. Taken together, these cases support the core claim of the paper: structural retrieval helps not only by improving relevance, but by presenting agents with a more execution-ready context.

\end{document}